\documentclass[sigconf]{acmart}
\usepackage{subcaption}
\usepackage{amsmath} 
\DeclareMathOperator{\Corr}{Corr}
\DeclareMathOperator{\Cov}{Cov}
\DeclareMathOperator{\Var}{Var}
\DeclareMathOperator{\E}{\mathbb{E}} 
\usepackage{booktabs} % For nice tables
\usepackage{mathtools}
\usepackage{bm}       % Bold math
\usepackage{float}    % For figure placement
\usepackage[table]{xcolor}
\usepackage[ruled,vlined]{algorithm2e}
\usepackage{multirow}
\usepackage{wrapfig}
\usepackage{etoc}     % For local table of contents
\usepackage{placeins} % For \FloatBarrier
\usepackage{caption}
\usepackage{needspace}
\usepackage[table]{xcolor}
\usepackage{booktabs}
\usepackage{tabularx}
\newcolumntype{C}{>{\centering\arraybackslash}X}
\usepackage{siunitx}
\acmDOI{}
\acmISBN{}

\definecolor{cream}{RGB}{255,242,204}
\definecolor{rowgray}{gray}{0.95}

\DontPrintSemicolon

\DeclareMathOperator{\MSE}{MSE}

\usepackage{amsmath,amsfonts,bm}

\def\eqref#1{equation~\ref{#1}}
\def\1{\bm{1}}

\def\eps{{\epsilon}}

\def\vc{{\bm{c}}}

\def\vt{{\bm{t}}}
\def\vu{{\bm{u}}}
\def\vv{{\bm{v}}}
\def\vw{{\bm{w}}}

\def\vy{{\bm{y}}}
\def\vz{{\bm{z}}}

\def\mC{{\bm{C}}}

\def\mT{{\bm{T}}}

\DeclareMathAlphabet{\mathsfit}{\encodingdefault}{\sfdefault}{m}{sl}
\SetMathAlphabet{\mathsfit}{bold}{\encodingdefault}{\sfdefault}{bx}{n}

\newcommand{\Ls}{\mathcal{L}}
\newcommand{\R}{\mathbb{R}}

\newcommand{\softmax}{\mathrm{softmax}}

\begin{document}

\title{Peeling Context from Cause for Molecular Property Prediction}

\author{Tao Li}
\affiliation{%
  \department{Department of Computer Science}
  \institution{Emory University}
  \city{Atlanta}
  \state{GA}
  \postcode{30322}
  \country{United States}
}
\email{tli349@emory.edu}

\author{Kaiyuan Hou}
\affiliation{%
  \department{Department of Computer Science}
  \institution{Emory University}
  \city{Atlanta}
  \state{GA}
  \postcode{30322}
  \country{United States}
}
\email{khou6@emory.edu}

\author{Tuan Vinh}
\affiliation{%
  \department{Medical Sciences Division}
  \institution{University of Oxford}
  \city{Oxford}
  \postcode{OX1 3PA}
  \country{United Kingdom}
}
\email{tuan.vinh@hertford.ox.ac.uk}

\author{Monika Raj}
\affiliation{%
  \department{Chemistry}
  \institution{Emory University}
  \city{Atlanta}
  \state{GA}
  \postcode{30322}
  \country{United States}
}
\email{mraj4@emory.edu}

\author{Carl Yang}
\affiliation{%
  \department{Computer Science}
  \institution{Emory University}
  \city{Atlanta}
  \state{GA}
  \postcode{30322}
  \country{United States}
}
\email{jyang71@emory.edu}



\begin{abstract}
Deep models are used for molecular property prediction, yet they are often hard to interpret and may rely on spurious context rather than causal structure, degrading reliability under distribution shift and harming predictive performance. We introduce \textbf{CLaP}, \emph{Causal Layerwise Peeling}, a framework that separates causal signal from context in a layerwise manner and integrates diverse graph representations of molecules. At each layer, a \emph{causal block} performs a soft split into causal and context branches, fuses causal evidence across modalities, and \emph{peels} batch-coupled context to concentrate on label-relevant structure, limiting shortcut signals and stabilizing layerwise refinement. Across nine molecular benchmarks, including OOD settings, CLaP reliably reduces MAE and MSE relative to competitive baselines. Beyond predictive performance, we conduct targeted causal evaluations under controlled environment shifts. These analyses provide empirical evidence of clear separation between environment-dependent context and causal signals.
As further evidence, we also analyze atom-level causal saliency maps and counterfactual outcomes, which highlight property-relevant substructures and show consistent alignment with chemical intuition.

\end{abstract}


\begin{teaserfigure}
    \centering
    \includegraphics[width=.93\linewidth,
        height=0.33\textheight,
        ]{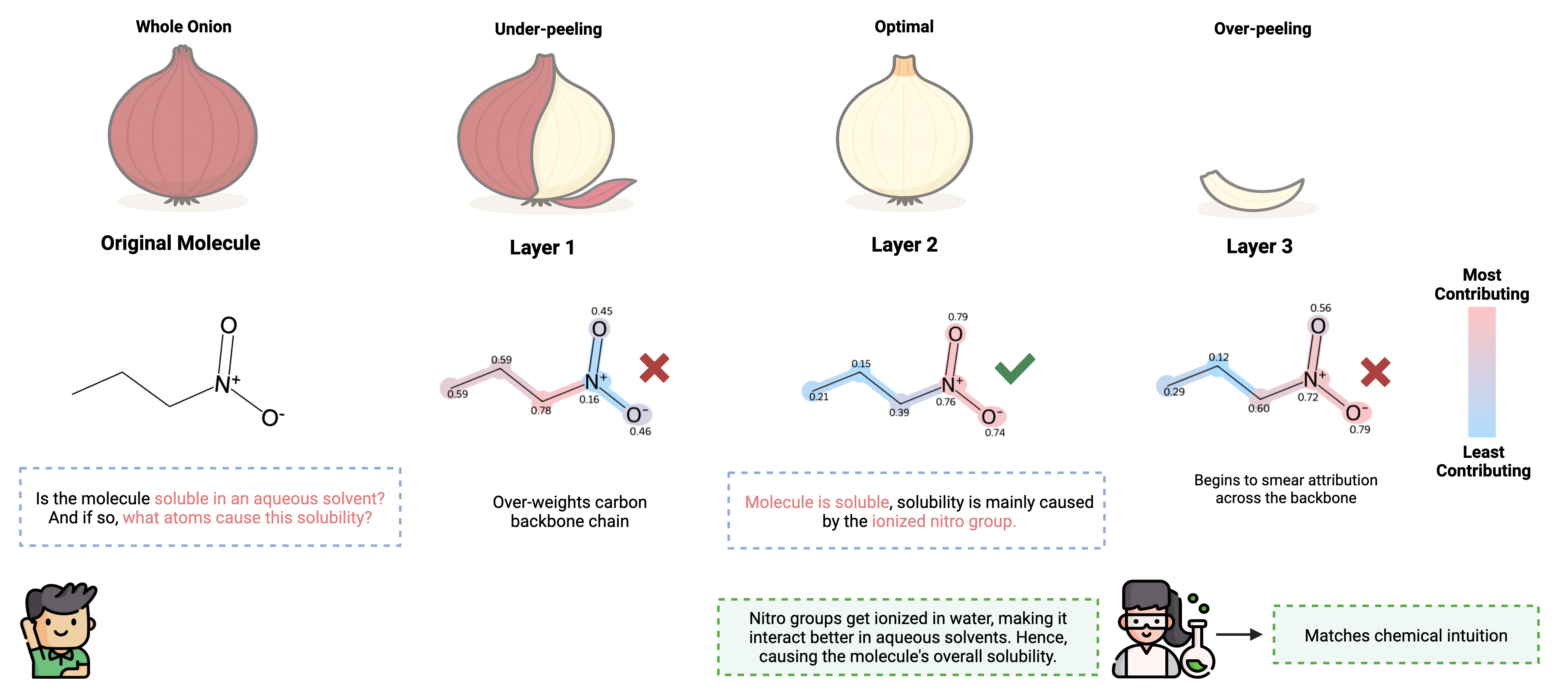}
    \Description{Illustration of layer-wise causal peeling on a molecule, showing attribution shifts from peripheral groups to the backbone across layers.}
    \caption{The original molecule in this illustrative example is iteratively “peeled” across layers. At shallow layers (layer 1), attribution underpeels the carbon backbone. At the optimal layer, causal weights concentrate on the nitro group, yielding chemically plausible solubility drivers. Deeper peeling begins to smear attribution across the backbone, reflecting over-peeling.}
    \label{fig:onion}
\end{teaserfigure}

\maketitle

\section{Introduction}
Designing molecules with desired properties is a central goal in drug discovery and materials design \citep{SanchezLengeling2018Inverse}. Graph-based deep learning is effective for property prediction \citep{Wu2018MoleculeNet,goodfellow2016deep}. However, models often inherently exploit spurious correlations tied to datasets or batches \citep{Geirhos2020Shortcut}, which hurts reliability under distribution shift \cite{arjovsky2019irm, li2025ood_graph_survey, li24s, an2025csib}. They also provide limited interpretable substructure-level guidance \citep{JimenezLuna2020XAI}, reducing their value for design. These gaps motivate the development of predictors that separate causal signal from contextual shortcuts and yield chemically meaningful attributions~\citep{CAL, CGL}.

\begin{figure*}[ht]
\centering
\includegraphics[width=2\columnwidth]{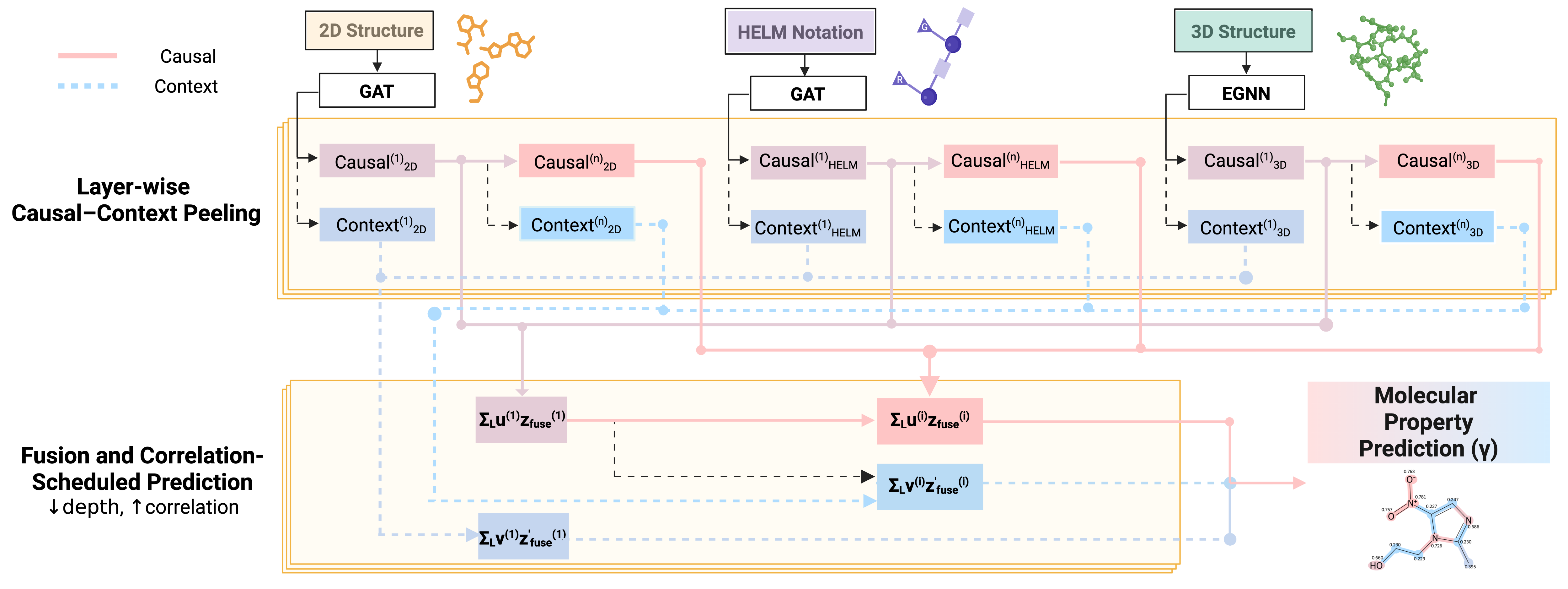}
\caption{Overview of the causal layerwise peeling architecture for \(L=2\) layers.}
\label{fig:model_architecture}
\end{figure*}

Prior work on invariant rationales promotes causal subgraphs by enforcing invariance across \emph{constructed} environments \citep{GraphOOD,wang2024dissecting,DIR,CAL,CIGA, peters2016invariantprediction}. While influential, these approaches are largely designed for classification, rely on synthetic environments whose fidelity to real data is uncertain, and often treat causality as a one-shot selection problem. We take a different route and introduce \textbf{CLaP}, \emph{Causal Layerwise Peeling}, a layerwise framework that peels context from cause. At each layer, a learnable splitter routes features into a \emph{causal} branch and a \emph{context} branch, progressively \emph{peeling} batch-coupled context and concentrating on label-relevant structure. Rather than engineering synthetic environments such as varying graph perturbations, or selecting a sparse subgraph with a predictor trained to be stable across them, we exploit the natural fluctuations of mini-batches as a source of contextual variation, which the model explicitly separates and absorbs through a dedicated context branch.

Our design is grounded in a \emph{batch-wise invariance} principle. As batch composition changes during training, only sample intrinsic signal should keep a stable alignment with the label. We implement this with a depth-dependent correlation target and a monotonicity regularizer. The causal readout increases its within-batch Pearson correlation with the label from shallow to deep layers. A residual objective trains the context branch on the remaining error. This enforces a clean division of labor and prevents shortcut leakage into the causal path. Building on this premise, we formulate a \emph{regression-ready} objective that aligns the scalar causal readout with continuous labels and enforces non-decreasing correlation across depth. These designs enable CLaP to \emph{peel} context from cause without synthetic environments and to provide robust, interpretable predictions.

From a molecular-property standpoint, different representations usually capture complementary facets of the same molecule~\citep{xu2024graph, zhang2024mvmrl, luo2025molcl_sp}. We therefore fuse causal evidence from three views—2D topology, peptide notation~\citep{Zhang2012HELM}, and 3D geometry~\citep{schutt2017schnet}. A gating module assigns sample-specific weights, up-weighting the modality that carries batch-invariant signal at each layer. Gating weights yield atom-level causal saliency maps, and layerwise fusion peels away context to reveal label-driving substructures (see Fig.~\ref{fig:onion}).

Experiments on small-molecule benchmarks (ESOL, FreeSolv, Lipo)~\citep{ESOL,FreeSolv,Lipo} and cyclic peptides (CycPeptMPDB)~\citep{CycPeptMPDB}, as well as OOD evaluations on Half\_Life\_Obach, Solubility, LD50\_Zhu, Hepatocyte, and Microsome~\citep{tdc}, show that CLaP achieves the most consistent improvements in MAE and MSE over both architecture-centric baselines (MAT, FP\text{-}GNN, MolFCL, GSL)~\citep{MAT,Cai2022FPGNN,MolFCL, GSL} and causality-oriented baselines (DIR, CAL, CGR)~\citep{DIR,CAL,CGL}. Beyond predictive improvements, we explicitly evaluate whether the learned representations exhibit the intended causal behavior.
We design controlled environment shifts at test time, including structural aggregation and re-batching interventions, to examine how the causal and context branches respond.
These evaluations indicate that CLaP tends to isolate signals that remain stable across different environments.
In addition, atom-level causal saliency maps naturally induced by the model’s causal gating mechanism are consistent with known structure–property relationships and, through case studies and counterfactual analyses, help identify substructures that are likely to causally influence the target properties.

In conclusion, our contributions are as follows:

\begin{itemize}
  \item
We propose CLaP, a multimodal, layerwise framework that separates \emph{causal} and \emph{context} representations and fuses causal evidence across molecular views, achieving empirically consistent performance improvements over strong baselines while improving prediction accuracy and robustness.

  \item
We evaluate the causal behavior of learned representations under controlled environment shifts, including scaffold-level aggregation and re-batching, to examine signal stability and the separation between causal and context branches.

  \item
CLaP yields atom-level causal saliency maps induced by its gating mechanism that highlight property-relevant substructures consistent with structure--property relationships.

\end{itemize}

\section{Methodology}

We first formalize a \emph{batch-wise invariance} principle that provides the theoretical basis for our supervised causal objective. We then instantiate it with a layerwise causal--context architecture.

\subsection{Batch-wise Invariance}
\label{sec:batch-invariance}

During training, each molecule is repeatedly sampled into different mini-batches across epochs, and thus encounters varying \emph{batch contexts} defined by co-occurring samples and label statistics.
A reliable predictor should maintain consistent alignment with the ground-truth label regardless of such batch fluctuations.
This observation motivates a \emph{batch-wise invariance} principle: only a sample-intrinsic signal sustains stable correlation with labels under batch contexts.

\paragraph{Setup.}
Index mini-batches by \(e \in \mathcal{E}\) over random shuffles.
Assume an additive view of the prediction \(s(x, e)\)
where \(c(x)\) is a sample-intrinsic component.
Let \(q(x,e)\) denote \emph{context} features whose association with \(y\) may depend on the batch \(e\), and consider a linearization of the layer’s scalar predictor with coefficients \(a,b\).

\begin{equation}
s(x,e) \;=\; a\, c(x) \;+\; b\, q(x,e),
\label{eq:bi-pred}
\end{equation}

\paragraph{Invariance objective.}
For any batch \(e\) of size \(B_e\), define batch-centered variables ${s}_e$ and the within-batch Pearson correlation. Batch-wise invariance requires 
% \[
% \widetilde{u}_e \;=\; u \;-\; \tfrac{1}{B_e}(\mathbf{1}^\top u)\,\mathbf{1},
% \]

% \begin{equation}
% \mathrm{Corr}_e(s,y)
% \;=\;
% \frac{\langle \widetilde{s}_e,\,\widetilde{y}_e\rangle}
%      {\|\widetilde{s}_e\|_2\,\|\widetilde{y}_e\|_2 + \varepsilon},
% \qquad \varepsilon>0\ \text{small}.
% \label{eq:bi-corr}
% \end{equation}

\begin{equation}
\mathrm{Corr}_e\!\big(s(x,e),\,y\big) \;\approx\; \rho
\quad \text{for all } e \in \mathcal{E},\ \text{fixed}\ \rho\in[-1,1]\ .
\label{eq:bi-inv}
\end{equation}

i.e., small dispersion of \(\mathrm{Corr}_e(s,y)\) across changing batches.

% A convenient surrogate is
\begin{equation}
\min_{a,b}\; \sum_{e \in \mathcal{E}}\Big(\mathrm{Corr}_e\!\big(s(x,e),y\big)-\rho\Big)^{2}.
\label{eq:bi-sur}
\end{equation}

\paragraph{Assumption.}
We make the following assumptions:
\begin{itemize}
\item[A1.] \(\mathrm{Corr}_e\!\big(c(x),y\big)=\kappa\) is constant in \(e\) (sample-intrinsic).
\item[A2.] \(\mathrm{Corr}_e\!\big(q(x,e),y\big)\) varies with \(e\) (batch-sensitive context).
\item[A3.] \(\mathrm{Var}_e\!\big(c(x)\big)\) and \(\mathrm{Var}_e\!\big(q(x,e)\big)\) are bounded away from \(0\).
\end{itemize}
Then any predictor of the form \eqref{eq:bi-pred} that satisfies the batch-wise invariance requirement \eqref{eq:bi-inv} for all \(e \in \mathcal{E}\) must necessarily have \(b=0\).
Equivalently, the batch-invariant solution thus only discards \(q\) and retains \(c\), and \(\mathrm{Corr}_e(s,y)=\mathrm{Corr}_e(c,y)=\kappa\) for all \(e\).
A proof for invariance is provided in Appendix~\ref{app:bi-analysis}.

\subsection{Layer-wise Causal–context Peeling}
\label{sec:training}

Our framework integrates three molecular views—2D graphs (SM), HELM notations (PE), and 3D geometries (GE). At each layer, causal blocks softly split features into \emph{causal} and \emph{context} paths. Causal signals are fused across modalities and aligned with the label via a depth-wise correlation schedule, while residual context is peeled and accumulated with stop-gradient. The final prediction combines the last causal readout and the accumulated residual (see Fig.~\ref{fig:model_architecture}).

\paragraph{Causal block.}
From the modality-specific node embeddings produced by the backbone encoders, each block at depth $\ell\!\in\!\{1,\dots,L\}$ applies a learnable node-wise splitter that assigns every atom/group a causal gate $\alpha_c^{(\ell)}\!\in\![0,1]$. Its complement $\alpha_t^{(\ell)}\!=\!1-\alpha_c^{(\ell)}$ acts as the context gate. These gates induce two branches per modality: a causal representation $\vz_c^{(\ell)}$ and a context representation $\vz_t^{(\ell)}$. For each modality at depth $\ell$, we obtain causal vectors $\vz_{\textit{SM}}^{(\ell)},\ \vz_{\textit{PE}}^{(\ell)},\ \vz_{\textit{GE}}^{(\ell)}\!\in\!\R^{D}$ from the current block, along with the context vectors $\vz_{\textit{SM}}^{\prime(\ell)},\ \vz_{\textit{PE}}^{\prime(\ell)},\\
$$ 
\vz_{\textit{GE}}^{\prime(\ell)}\!\in\!\R^{D}$. A gating network $g_F(\cdot)$ produces sample-specific soft weights $w^{(\ell)}$ over modalities via a softmax, and these weights are used to form the fused causal representation $\vz_{\textit{fuse}}^{(\ell)}$ (and the fused context representation $\mathbf{z}_{\mathrm{\textit{fuse}}}^{\prime(\ell)}$) at depth $\ell$.
\[
\vw^{(\ell)}=\softmax\!\Big(g_F\big([\vz^{(\ell)}_{\textit{SM}};\vz^{(\ell)}_{\textit{PE}};\vz^{(\ell)}_{\textit{GE}}]\big)/\tau\Big)\in\R^{3},
\]

Scalar readouts use projection heads. Applying causal head $\vu^{(\ell)}$ to $\mathbf{z}_{\textit{fuse}}^{(\ell)}$ yields $\vc^{(\ell)}$, while applying context head $\vv^{(\ell)}$ to $\mathbf{z}_{\textit{fuse}}^{\prime(\ell)}$ yields $\vt^{(\ell)}$. The causal branch is forwarded to depth $\ell{+}1$ (yielding $\vc^{(\ell+1)}$) while an additional context component $t^{(\ell+1)}$ is peeled. This recursion continues until depth $L$. After block \(L\), the per-batch scalars are stacked across layers to enable supervision and analysis.

\[
\mC=\big[\vc^{(1)}\!\cdots\!\vc^{(L)}\big],\qquad
\mT=\big[\vt^{(1)}\!\cdots\!\vt^{(L)}\big]\ \in\R^{B\times L}.
\]

\paragraph{Causal branch: depth-guided correlation.}
The causal branch is designed to focus, layerwise, on the label-predictive signal. 
At each depth $\ell$, we align the fused causal scalar $\vc^{(\ell)}$ with the batch labels $\vy\!\in\!\R^{B}$ by \emph{explicitly} constraining their Pearson correlation to follow a depth-dependent target schedule in a supervised manner.

% We first center the variables and then calculate the mini-batch Pearson correlation at depth $\ell$ with a small stabilizer $\eps>0$.

\begin{equation}
\begin{gathered}
% \begin{aligned}
% \widetilde{\vc}^{(\ell)} &= \vc^{(\ell)} - \tfrac{1}{B}(\vone^\top \vc^{(\ell)})\,\vone \qquad
% \widetilde{\vy} &= \vy - \tfrac{1}{B}(\vone^\top \vy)\,\vone,
% \end{aligned}\\[2pt]
\mathrm{corr}_\ell = \frac{\langle \vc^{(\ell)},\,{\vy}\rangle}
{\|{\vc}^{(\ell)}\|_2\,\|{\vy}\|_2+\eps} \in [-1,1].
\end{gathered}
\label{eq:corr}
\end{equation}

We set a linearly increasing target correlation schedule $\rho_\ell$ from $\rho_{\min}$ to $\rho_{\max}$ so that deeper layers are encouraged to capture progressively finer, label-critical variation. Then penalize deviations between the batchwise Pearson correlation and its target,
\begin{align}
\Ls_{\mathrm{corr}}
= \frac{1}{L}\sum_{\ell=1}^{L}\big(\mathrm{corr}_\ell - \rho_\ell\big)^{2},
\label{eq:Lcorr}
\end{align}
The correlation loss guides each layer toward the target depth-wise trend and concentrates causal signal while filtering out irrelevant batch-specific context noise in a principled way.

To stabilize the hierarchy and ensure consistent refinement of causal regression from $\ell$ to $\ell{+}1$, we further apply a monotonicity regularizer, encouraging correlations to be non-decreasing from layer to layer via a hinge penalty with margin $\gamma\!\ge\!0$:
\begin{align}
\Ls_{\mathrm{mono}}
= \frac{1}{L-1}\sum_{\ell=1}^{L-1}\max\!\Big\{0,\ \mathrm{corr}_\ell - \mathrm{corr}_{\ell+1} + \gamma \Big\}.
\label{eq:Lmono}
\end{align}

Together, \eqref{eq:Lcorr}–\eqref{eq:Lmono} encourage a depth-wise progression of representation. 
Shallow layers capture coarse, high-variance patterns, and deeper layers increasingly model causal patterns.

\begin{table*}[h]
\centering
\caption{Comparison across nine benchmarks and ablation results. Best per-dataset/metric in \textbf{bold}. Lower is better.}
\label{tab:method_cmp}
\setlength{\tabcolsep}{1pt}
\renewcommand{\arraystretch}{1.3}
\resizebox{1.0\linewidth}{!}{
% {\fontsize{11pt}{13pt}\selectfont
% \fontseries{l}\selectfont
\begin{tabular}{l*{9}{cc}}
\toprule

% \rowcolor{headgray}
\textbf{Method} &
\multicolumn{2}{c}{\textit{\textbf{ESOL}}} &
\multicolumn{2}{c}{\textit{\textbf{FreeSolv}}} &
\multicolumn{2}{c}{\textit{\textbf{Lipo}}} &
\multicolumn{2}{c}{\textit{\textbf{CycPeptMPDB}}} &
\multicolumn{2}{c}{\textit{\textbf{Half Life}}} &
\multicolumn{2}{c}{\textit{\textbf{Hepatocyte}}} &
\multicolumn{2}{c}{\textit{\textbf{Microsome}}} &
\multicolumn{2}{c}{\textit{\textbf{Solubility}}} &
\multicolumn{2}{c}{\textit{\textbf{Toxicity}}} \\
% \rowcolor{headgray}
\multicolumn{19}{c}{}\\[-1.6ex]
\cmidrule(r){2-3}
\cmidrule(lr){4-5}
\cmidrule(lr){6-7}
\cmidrule(lr){8-9}
\cmidrule(lr){10-11}
\cmidrule(lr){12-13}
\cmidrule(lr){14-15}
\cmidrule(lr){16-17}
\cmidrule(l){18-19}

% \rowcolor{headgray}
& MAE & MSE & MAE & MSE & MAE & MSE & MAE & MSE &
  MAE & MSE & MAE & MSE & MAE & MSE & MAE & MSE & MAE & MSE \\
\midrule

% ---------- Row 1: FP-GNN (grey) ----------
\rowcolor{gray!15}
\textbf{FP-GNN} &
0.5039 & 0.5150 &
1.1561 & 2.0783 &
0.4809 & 0.4890 &
0.3487 & 0.2202 &
0.8857 & 1.5464 &
1.0836 & 1.5861 &
0.8083 & 1.2308 &
0.8383 & 1.2841 &
0.5242 & 0.5100 \\

% ---------- Row 2: GSL (grey) ----------
\textbf{GSL} &
1.0032 & 1.2566 &
2.3476 & 4.0343 &
0.9007 & 1.1478 &
0.6590 & 0.8127 &
0.9365 & 1.3304 &
1.1045 & \textbf{\normalsize 1.3201} &
1.0723 & 1.2428 &
1.5883 & 1.9713 &
0.7414 & 0.9073 \\

% ---------- Row 3: MAT (grey) ----------
\rowcolor{gray!15}
\textbf{MAT} &
0.5394 & 0.4803 &
0.7680 & 1.2255 &
0.5284 & 0.4745 &
0.4129 & 0.3146 &
0.9915 & 1.5690 &
1.5660 & 2.6050 &
1.1350 & 1.5820 &
0.9250 & 1.2220 &
0.8622 & 0.9300 \\

% ---------- Row 4: CAL (white) ----------
\rowcolor{gray!15}
\textbf{CAL} &
0.4923 & 0.4155 &
1.1273 & 2.2452 &
0.6217 & 0.6536 &
0.3682 & 0.2202 &
1.0189 & 1.6643 &
0.9839 & 1.6163 &
0.8335 & 1.4078 &
1.0572 & 1.9931 &
0.6636 & 0.7446 \\

% ---------- Row 5: CGR (white) ----------
\textbf{CGR} &
1.5853 & 4.0291 &
2.7268 & 11.4813 &
0.9670 & 1.5269 &
0.5792 & 0.5578 &
1.0360 & 1.6673 &
1.2371 & 1.9796 &
1.2544 & 1.9550 &
1.6741 & 4.5156 &
0.7794 & 1.0112 \\

% ---------- Row 6: DIR (white) ----------
\rowcolor{gray!15}
\textbf{DIR} &
0.6267 & 0.6794 &
2.2420 & 7.8170 &
0.6328 & 0.7313 &
0.4243 & 0.2927 &
0.9600 & 1.6880 &
1.0780 & 1.8210 &
1.1580 & 2.1250 &
1.1890 & 4.5220 &
0.6810 & 0.8050 \\

% ---------- Row 7: MolFCL (white) ----------
\textbf{MolFCL} &
0.4660 & 0.4740 &
1.0300 & 1.9890 &
\textbf{\normalsize 0.4330} & 0.3670 &
0.3690 & 0.2440 &
0.9210 & 1.5000 &
1.1160 & 1.7660 &
0.9430 & 1.3580 &
\textbf{\normalsize 0.6990} & \textbf{\normalsize 0.9060} &
0.7530 & 1.0920 \\

\midrule

\rowcolor{gray!15}
w/o split &
0.5094 & 0.4939 &
1.2858 & 2.5719 &
0.5145 & 0.4302 &
0.3120 & 0.1834 &
0.8301 & 1.2055 &
1.0571 & 1.6711 &
0.7908 & 1.1071 &
0.7850 & 1.1157 &
0.5184 & 0.5090 \\

w/o correlation &
0.5078 & 0.4767 &
0.8779 & 1.3742 &
0.5562 & 0.5122 &
0.3215 & 0.1866 &
0.8222 & \textbf{\normalsize 1.1382} &
1.0230 & 1.5725 &
0.8051 & 1.0930 &
0.7835 & 1.1295 &
0.5250 & 0.5087 \\

\rowcolor{gray!15}
w/o context &
0.4732 & 0.4137 &
1.0753 & 1.7338 &
0.5824 & 0.5348 &
0.3214 & 0.2037 &
0.8123 & 1.1543 &
1.0330 & 1.5965 &
0.7805 & 1.0647 &
0.7801 & 1.1035 &
0.5189 & 0.4993 \\

% ---------- Row 8: CLaP (highlighted) ----------
% \rowcolor{gray!25}
\rowcolor{cream}
\textbf{CLaP} &
\textbf{\normalsize 0.4456} & \textbf{\normalsize 0.3583} &
\textbf{\normalsize 0.7020} & \textbf{\normalsize 0.8866} &
0.4672 & \textbf{\normalsize 0.3645} &
\textbf{\normalsize 0.3056} & \textbf{\normalsize 0.1644} &
\textbf{\normalsize 0.7832} & 1.1537 &
\textbf{\normalsize 0.9520} & 1.4108 &
\textbf{\normalsize 0.7221} & \textbf{\normalsize 0.9682} &
0.7725 & 1.0919 &
\textbf{\normalsize 0.5064} & \textbf{\normalsize 0.4858} \\

\bottomrule
\end{tabular}
}
\end{table*}

\paragraph{Context branch: context learning.}
The context branch isolates environment- and batch-dependent variation that may contribute to prediction but is not stable across environments in practice. By removing these factors from the causal branch, the causal signal becomes clearer and more discriminative. At depth $\ell$, each block yields a context scalar $\vt^{(\ell)}$. We aggregate these across layers, while retaining only the final causal scalar as the causal output:
\begin{align}
t = \sum_{\ell=1}^{L} \vt^{(\ell)} \in \R^{B}, \qquad
y_{\mathrm{c}}^\star = \vc^{(L)}.
\label{eq:t-sum}
\end{align}
To maintain a clear division of labor, the context branch is trained on the residual between label $\vy$ and final causal output, with gradients stopped through $y_{\mathrm{c}}^\star$ to avoid contaminating the causal pathway:
\begin{align}
\Ls_{\mathrm{ctx}}
= \mathrm{MSE}\!\Big(t,\ \vy - \mathrm{stopgrad}(y_{\mathrm{c}}^\star)\Big).
\label{eq:Lctx}
\end{align}
This design assigns the stable, generalizable component to the causal path, while the context path captures the remaining residual variation, calibrating $y_{\mathrm{c}}^\star$ by absorbing context-specific noise, such as molecular scaffolds or batch-level statistical effects.

% to causal branch -> to the causal branch

\paragraph{Prediction and total objective.}
The model predicts by additively combining depth-$L$ causal readout with the accumulated context:
\begin{equation}
\widehat{y}=y_{\mathrm{c}}^\star + t,\qquad 
\Ls_{\mathrm{pred}}=\MSE(\widehat{y},\,\vy).
\label{eq:pred}
\end{equation}

We regularize the causal branch toward \emph{batch-invariant}, depth-increasing alignment with the label, while training the context branch on the residual. The overall objective function integrates these structural priors into a joint loss formulated as:
% \[
% \begin{aligned}
% \mathcal{L}_{\mathrm{total}} = 
%     \mathcal{L}_{\mathrm{pred}} &+ \underbrace{\lambda_{\mathrm{caus}}\,\mathcal{L}_{\mathrm{corr}}}_{\text{causal alignment}} \\
%     &+ \underbrace{\lambda_{\mathrm{mono}}\,\mathcal{L}_{\mathrm{mono}}}_{\text{depth monotonicity}} + \underbrace{\lambda_{\mathrm{unif}}\,\mathcal{L}_{\mathrm{ctx}}}_{\text{residual fit}}
% \end{aligned}
% \]
% \begin{equation}
%     \mathcal{L}_{\mathrm{total}} = \mathcal{L}_{\mathrm{task}} + \lambda_{\mathrm{caus}}\mathcal{L}_{\mathrm{inv}} + \lambda_{\mathrm{mono}}\mathcal{L}_{\mathrm{mono}} + \lambda_{\mathrm{unif}}\mathcal{L}_{\mathrm{unif}}
% \end{equation}
\[
\begin{split}
\Ls_{\mathrm{total}}
= \Ls_{\mathrm{pred}}
+ \underbrace{\lambda_{\mathrm{caus}}\,\Ls_{\mathrm{corr}}}_{\text{causal alignment}}
+ \underbrace{\lambda_{\mathrm{mono}}\,\Ls_{\mathrm{mono}}}_{\text{depth monotonicity}}
+ \underbrace{\lambda_{\mathrm{unif}}\,\Ls_{\mathrm{ctx}}}_{\text{residual fit}}, \\
\end{split}
% \]
% \[
% \lambda_{\mathrm{caus}},\lambda_{\mathrm{mono}},\lambda_{\mathrm{unif}}\ge 0.
\]

\section{Experiments}

\subsection{Experimental Setup}

\noindent\textbf{Datasets.}
We evaluate CLaP on nine molecular property prediction datasets, categorized into two standardized evaluation settings: in-distribution (ID) and out-of-distribution (OOD).
An overview of dataset statistics is provided in Table~\ref{tab:our_dataset_stats}.

In the in-distribution setting, we apply random splits to four representative small-molecule and peptide benchmarks commonly used in prior work: ESOL (aqueous solubility)~\citep{ESOL}, FreeSolv (hydration free energy)~\citep{FreeSolv}, Lipo (octanol/water partition coefficient)~\citep{Lipo}, and CycPeptMPDB (passive permeability of cyclic peptides)~\citep{CycPeptMPDB}.

To evaluate robustness under distribution shift, we adopt preset scaffold-based OOD splits on five additional datasets from TDC~\citep{tdc}: Half\_Life\_Obach, Solubility, Toxicity, Hepatocyte, and Microsome, which cover absorption, excretion, and toxicity properties.

\noindent\textbf{Baselines.}
We compare CLaP with two major categories of molecular property prediction methods for direct comparison:
(i) \emph{architecture-centric} models that emphasize network design, including representative MAT~\citep{MAT} (Molecule Attention Transformer), FP-GNN~\citep{Cai2022FPGNN} (Fingerprint-enhanced GNN), GSL~\citep{GSL} (Graph Structure Learning), and MolFCL~\citep{MolFCL} (fragment- and functional-group-aware contrastive learning); and
(ii) \emph{causality-oriented} approaches that improve prediction by modeling causal structure or mitigating confounding effects, including DIR~\citep{DIR} (Discovering Invariant Rationales) and CAL~\citep{CAL} (Causal Attention Learning). We also include CGR~\citep{CGL} (Causal Graph Learning), which leverages causal intervention to enhance molecular regression. For causality-based models originally proposed for classification, we preserve their causal mechanisms and adapt only the output layer for regression. See Appendix~\ref{sec:baseline_implementation_details} for further baseline implementation details.

As summarized in Table~\ref{tab:method_comparison}, the considered baselines primarily operate on single- or dual-modality inputs, focusing on 2D molecular graphs alone or augmenting them with auxiliary representations such as molecular fingerprints or 3D structures. In contrast, CLaP jointly incorporates 2D graphs, 3D structural information, and optional sequence-level HELM representations for peptides within a unified framework. From a computational perspective, the additional cost introduced by CLaP is modest. It scales linearly with the number of nodes and modalities and remains a minor overhead compared to backbone message passing in practice, since the numbers of modalities and peeling layers are small and fixed. Moreover, compared to other causality-oriented approaches that rely on manually fabricated environment interventions, CLaP avoids additional computation associated with repeated intervention-based training.

\begin{table}[t]
\centering
\footnotesize
\caption{
Overview of datasets and method characteristics used in this study.
(\textbf{Top}) Dataset statistics under in-distribution (ID) and out-of-distribution (OOD) settings.
(\textbf{Bottom}) Comparison of computational complexity and input modalities.
}
\label{tab:dataset_and_method_summary}

\setlength{\tabcolsep}{5pt}
\renewcommand{\arraystretch}{0.9}

% ---------- Subtable (a): Dataset statistics ----------
\begin{subtable}[t]{\columnwidth}
\centering
\caption{Dataset statistics under ID and OOD settings.}
\label{tab:our_dataset_stats}

\begin{tabularx}{\columnwidth}{
    c
    >{\raggedright\arraybackslash}X
    c
    S[table-format=5.0]
    S[table-format=2.4]
}
\toprule
\textbf{Setting} & \textbf{Dataset} & \textbf{Type} &
{\textbf{\# Samples}} & {\textbf{Avg. \# Atoms}} \\
\midrule
\multirow{4}{*}{ID}
& ESOL     & Physical & 1128 & 13.2899 \\
& FreeSolv & Physical & 642  & 8.7227 \\
& Lipo     & Physical & 4171 & 27.0269 \\
& CycPeptMPDB & Physical & 6960 & 64.8966 \\
\midrule
\multirow{5}{*}{OOD}
& Half Life   & PK/ADME & 667  & 27.6977 \\
& Hepatocyte & PK/ADME & 1213 & 28.5812 \\
& Microsome  & PK/ADME & 1102 & 28.4519 \\
& Solubility & Physical & 9982 & 17.3830 \\
& Toxicity   & Safety   & 7385 & 16.1641 \\
\bottomrule
\end{tabularx}
\end{subtable}

\vspace{2pt}

% ---------- Subtable (b): Method comparison ----------
\begin{subtable}[t]{\columnwidth}
\centering
\caption{Computational complexity and input modalities across methods.}
\label{tab:method_comparison}

\begin{tabularx}{\columnwidth}{@{}l >{\raggedright\arraybackslash}X c@{}}
\toprule
\textbf{Method} & \textbf{Computational Complexity} & \textbf{Input Modality} \\
\midrule
FP-GNN & $O(L_{back}(V + E) + F)$ & 2D + FP \\
GSL    & $O(V^2 + L_{back}(V + E))$ & 2D + FP \\
MAT    & $O(L \cdot V^2)$ & 2D + 3D \\
CAL / CGR & $O(L_{back}(V + E) + B)$ & 2D \\
DIR    & $O(L_{back}(V + E) + |Env| \cdot B)$ & 2D \\
MolFCL & $O(L_{back}(V + E) + m^2)$ & 2D \\ 
\textbf{CLaP} & $\mathbf{O(L_{back}(V + E) + L \cdot M \cdot V)}$ & \textbf{2D + 3D + (HELM)} \\
\bottomrule
\end{tabularx}
\end{subtable}

\end{table}

\subsection{Performance Comparison}

Table~\ref{tab:method_cmp} compares CLaP against \emph{seven} baselines on \emph{nine} datasets spanning both ID (ESOL, FreeSolv, Lipo, CycPeptMPDB) and OOD (Half Life, Hepatocyte, Microsome, Solubility, Toxicity) settings. We report MAE and MSE. All results are averaged over 5 runs.

CLaP ranks first on \textbf{7/9} datasets in both MAE and MSE and yields substantial improvements over the strongest non--CLaP baselines.
On challenging ID benchmarks such as \textit{CycPeptMPDB} and \textit{FreeSolv}, CLaP achieves large gains,
and these improvements extend to OOD settings, including \textit{Microsome} and \textit{Toxicity}.
On the remaining tasks, CLaP remains competitive:
it attains the best MSE on \textit{Lipo}, achieves the lowest MAE on \textit{Hepatocyte} with near-best MSE,
while \textit{Solubility} is the only dataset where MolFCL leads on both metrics.

These results indicate that causal–context peeling—via the correlation schedule and residual context calibration—consistently improves data efficiency and robustness across both ID and OOD regimes.
This empirically supports the effectiveness of the proposed framework in disentangling and effectively leveraging causal and contextual information for molecular property prediction.

\subsection{Variants and Ablations}

\subsubsection{Framework design}

Table~\ref{tab:method_cmp} also reports ablations of framework components. 
The framework integrates depthwise correlation schedule (Eq.~\eqref{eq:Lcorr}) and the context branch (Eq.~\eqref{eq:Lctx}).

\noindent\textbf{w/o causal--context split.}
This variant removes the correlation loss $\Ls_{\mathrm{corr}}$, the context loss $\Ls_{\mathrm{c t x}}$, and the monotonicity penalty $\Ls_{\mathrm{mono}}$, thereby collapsing our two-branch design. Training reduces to plain regression with $\Ls_{\mathrm{pred}}$ (Eq.~\eqref{eq:pred}). Empirically, performance drops (e.g., on the CycPeptMPDB dataset), indicating that straightforward end-to-end fitting struggles to separate label-relevant structure and is easily distracted by spurious information.

Under the batch-wise invariance view, a predictor that mixes batch-dependent context \(q(x,e)\) with intrinsic signal \(c(x)\) yields \(\mathrm{Corr}_e(s,y)\) that fluctuates with \(e\).
Removing the split collapses the losses \(\Ls_{\mathrm{corr}},\Ls_{\mathrm{ctx}},\Ls_{\mathrm{mono}}\), allowing \(b\!\neq\!0\) in Eq.~\eqref{eq:bi-pred}, contradicting invariant solution driven by Eq.~\eqref{eq:bi-sur}.
The performance drop reflects increased reliance on batch-sensitive shortcuts.

\noindent\textbf{w/o correlation schedule.}
We omit the correlation loss \(\Ls_{\mathrm{corr}}\) and (Eq.~\eqref{eq:Lcorr}), which disables the target schedule for \(\rho\). The depth schedule \(\{\rho_\ell\}\) and \(\Ls_{\mathrm{corr}}\) (Eq.~\eqref{eq:Lcorr}) specify a curriculum from coarse to decisive label-relevant factors.

Without this schedule, the stack lacks directional guidance and drifts toward batch-specific shortcuts. It fits context patterns with the label signal, which hurts generalization. Keeping the monotonicity term \(\Ls_{\mathrm{mono}}\) is not enough because the non-decreasing constraint lacks direction and cannot steer which features to keep or discard. In the full model the schedule works as a curriculum that allocates capacity to increasingly predictive components and focuses deeper layers on label-relevant variation.

\noindent\textbf{w/o context branch.}
We ablate the context branch and predict solely from the final-layer causal output $y_{\mathrm{c}}^\star$. 
This eliminates residual calibration of environment-driven variation into prediction. See Appendix~\ref{app:Context_Branch_analysis} for a formal derivation and risk decomposition.

To illustrate, consider a molecule whose aqueous solubility is primarily driven by a polar functional group (e.g., a nitro or hydroxyl group), while the overall scaffold or ring system introduces secondary, context-dependent effects.
In this case, the causal prediction $y_{\mathrm{c}}^\star$ captures the contribution of the functional group but may deviate slightly from the ground-truth label $y$ due to scaffold-dependent or batch-specific factors.
The context branch learns to produce a residual term $y_{\text{ctx}} \approx y - y_{\mathrm{c}}^\star$, calibrating the final prediction as $\widehat{y} = y_{\mathrm{c}}^\star + y_{\text{ctx}}$.
Without this branch, chemical information related to scaffold- and batch-dependent effects cannot be incorporated to correct deviations in $y_{\mathrm{c}}^\star$, reducing overall predictive fidelity.

\begin{figure}[h]
  \centering
  \includegraphics[width=1.0\linewidth]{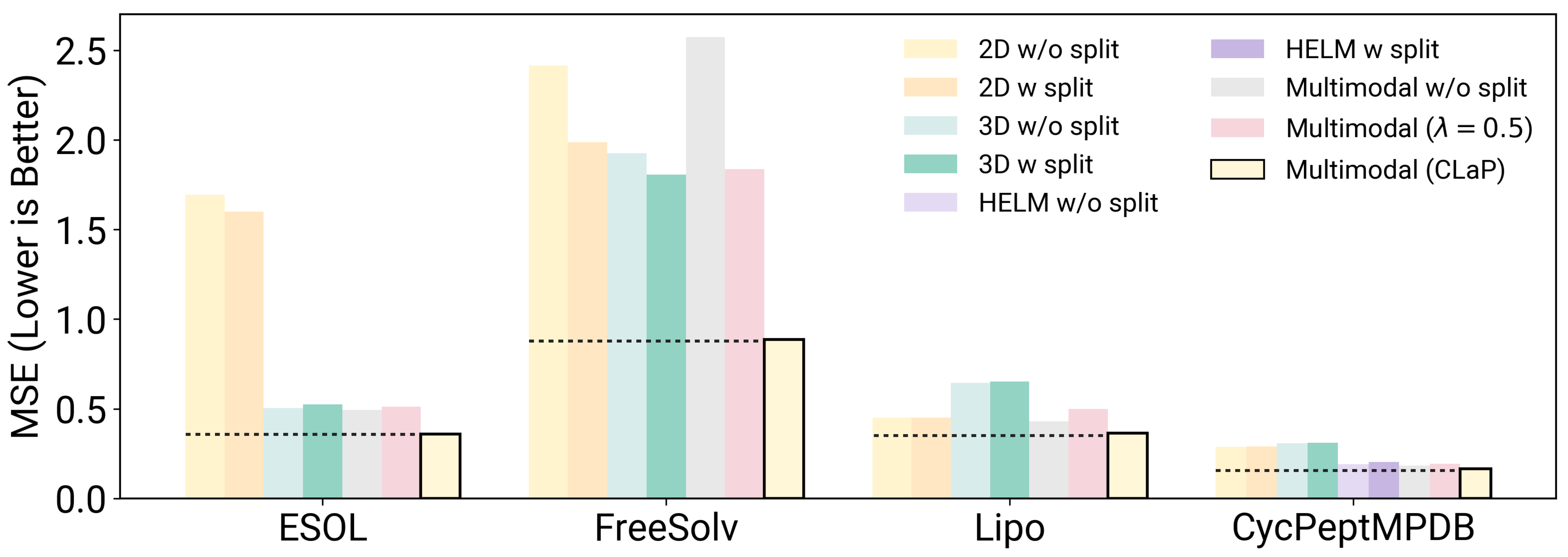}

\caption{Effect of causal peeling across modalities and consistency weight on prediction performance. Lower is better.}

  \label{tab:modalities_consistency}
\end{figure}

\subsubsection{Multi-modality Design}

In this section, we first examine how the proposed causal–context split behaves under different modality configurations. We consider both unimodal and multimodal settings, where CycPeptMPDB provides three complementary views per sample(SM, PE, GE), while ESOL, FreeSolv, and Lipo each provide two views (SM and GE). Fig.~\ref{tab:modalities_consistency} compares unimodal variants with multimodal fusion models, with and without the causal split. Across datasets with two or more modalities, enabling the causal--context split consistently improves both accuracy and stability over unimodal baselines. These results indicate that while multimodality supplies richer input signals, the causal--context split plays a role in structuring and exploiting multimodal information, allowing CLaP to leverage different modalities in a complementary manner.

\begin{figure*}[h]
  \centering
  \begin{subfigure}{0.45\linewidth}
    \centering
    \includegraphics[width=.9\linewidth,height=0.3\textheight,keepaspectratio]{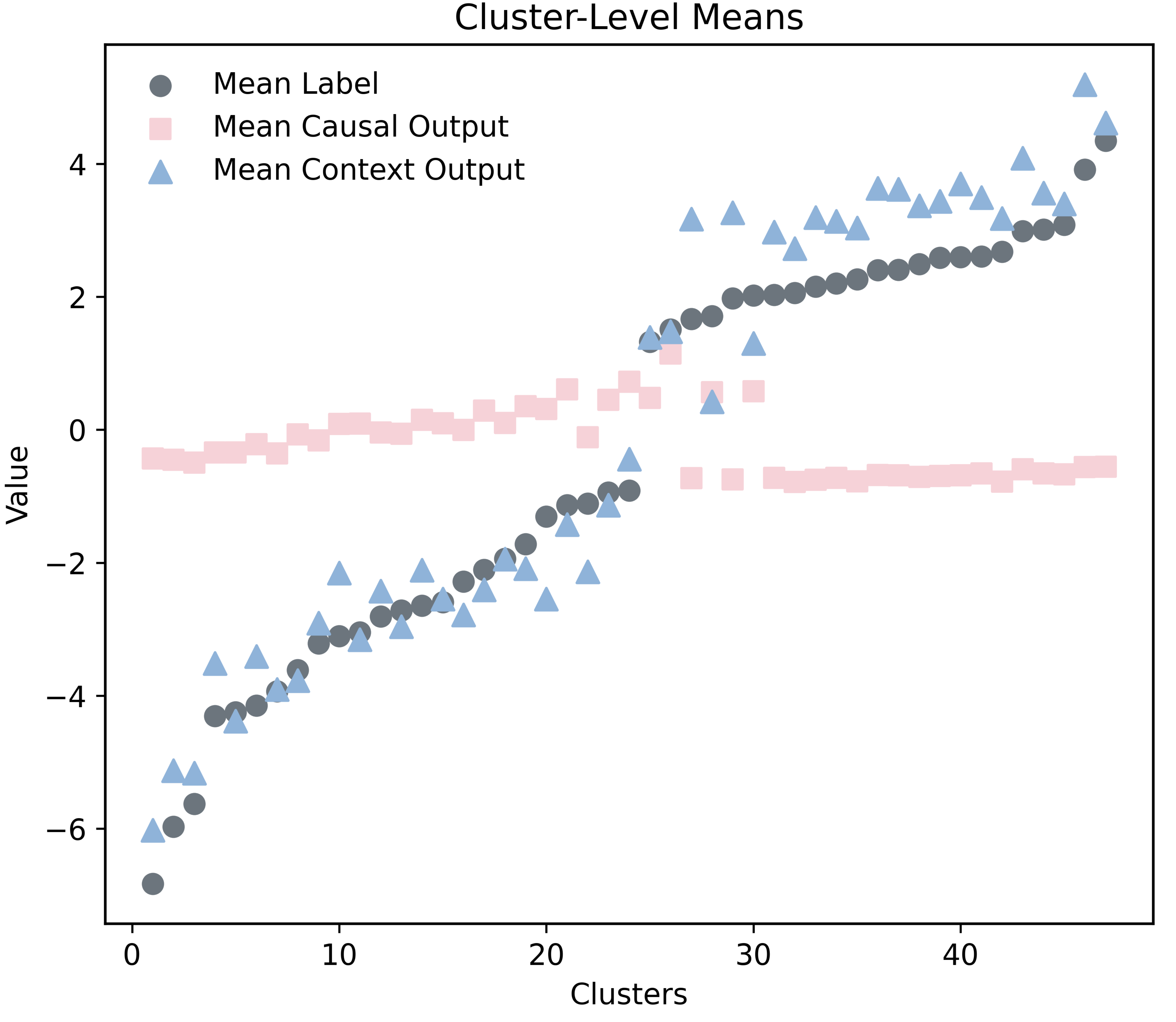}
    \caption{Causal analysis across molecular scaffolds.}
    \label{fig:causal_scaffolds}
  \end{subfigure}
  \hfill
  \begin{subfigure}{0.45\linewidth}
    \centering
    \includegraphics[width=.9\linewidth,height=0.3\textheight,keepaspectratio]{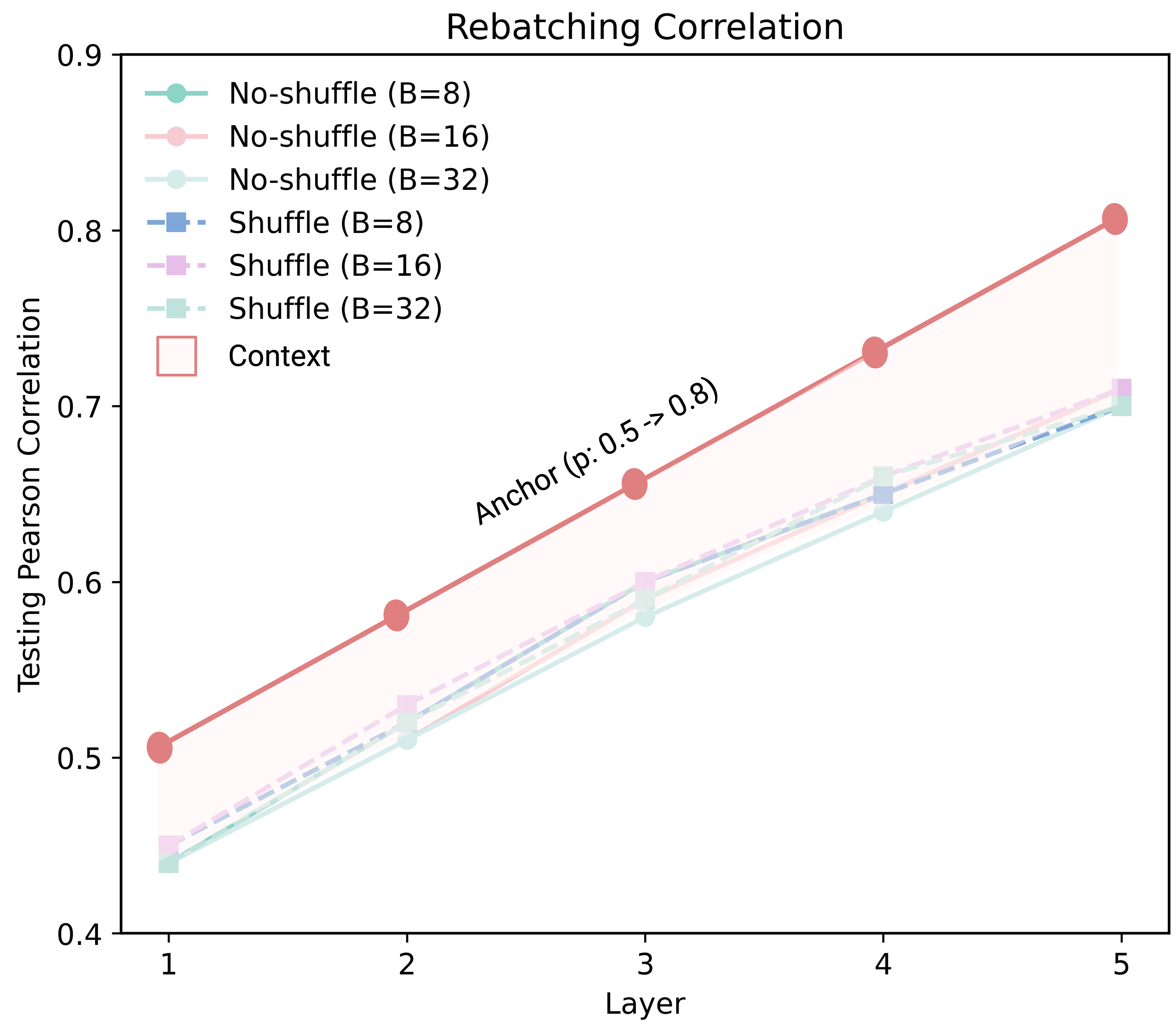}
    \caption{Environment interventions via re-batching.}
    \label{fig:causal_rebaching}
  \end{subfigure}

  \caption{Causal evaluation under scaffold and environment shifts.}
  \label{fig:causal_eval}
\end{figure*}

% In unimodal settings, the effect of causal peeling is limited and sometimes inconsistent. On \textit{ESOL} and \textit{Lipo}, both 2D and 3D variants exhibit only minor changes, and in some cases performance even degrades. For \textit{FreeSolv}, 2D and 3D improve slightly under causal splitting, but still remain far weaker than the multimodal fusion model. On \textit{CycPeptMPDB}, 2D, 3D, and HELM all gain marginally, yet consistently trail behind the fused configuration.

Next, we investigate whether explicitly enforcing cross-modal agreement is beneficial beyond the proposed causal split. To this end, we introduce a cosine-similarity penalty between modality-specific embeddings, weighted by $\lambda_{\text{cons}}$, and evaluate its effect (Fig.~\ref{tab:modalities_consistency}). Across datasets, adding this penalty degrades performance compared to the default setting without cross-modal consistency ($\lambda_{\text{cons}} = 0$), with larger values of $\lambda_{\text{cons}}$ leading to larger drops.

% The degradation is most substantial on FreeSolv and Lipo, and remains observable on CycPeptMPDB and ESOL.

Overall, the proposed split improves how information from multiple modalities is utilized effectively. By separating label-relevant signals from contextual residue, the model can more selectively leverage complementary cues across modalities. In contrast, without this separation, multimodal signals remain entangled, leading to weaker performance despite access to multiple inputs.

\subsection{Causal Evaluation under Shifts}

To examine whether CLaP exhibits intended causal behavior beyond predictive accuracy, we conduct evaluations under controlled environment shifts and analyze how the causal and context branches respond to changes in the environment at test time.

\subsubsection{Causal Effects across Structural Environments}

\paragraph{Evaluation setup.}
To evaluate whether the proposed CLaP can separate environment-level structure from sample-specific variation, we analyze its behavior from a structural hierarchy perspective.
The goal of this experiment is to examine whether the model can effectively disentangle shared scaffold-level effects from residual molecule-level variation across different structural scales.

We consider prediction results for all molecular properties in the test set.
For analytical clarity, we conceptually distinguish two sources of structure-related variation.
The first corresponds to scaffold structures defined by the Bemis--Murcko framework, which represent stable and shared environmental factors within each structural cluster.
The second consists of local structural variations attached to the scaffold, such as functional groups or substitution patterns, whose effects are typically expressed as molecule-specific differences.
This partition is introduced for causal analysis rather than as an assumption about underlying chemical mechanisms.

For each property, molecules are grouped into clusters according to their scaffolds, and only clusters containing at least five molecules are retained to ensure statistical stability.
Within each cluster, we compute the cluster-level means of the causal branch output, the context branch output, and the ground-truth property values.
This cluster-wise averaging operation statistically marginalizes molecule-level local variation, thereby isolating scaffold-level shared structural information in expectation across clusters.

To assess whether scaffold-induced systematic variation is correctly allocated across branches, we examine the correlations between cluster-level branch outputs and the aggregated labels.

\paragraph{Observed behavior.}

As shown in Fig.~\ref{fig:causal_scaffolds}, scaffold-induced baseline shifts are captured by the context branch, which shows strong correlation with the labels, whereas the causal branch remains almost invariant across scaffolds and exhibits weaker correlation.

% This behavior follows directly from the design of the scaffold-level aggregation.
This pattern arises from the scaffold-level aggregation used in this evaluation and provides direct evidence for the intended causal separation in our model.
By averaging predictions and labels within each scaffold cluster, molecule-specific variations—primarily functional-group-level effects—are statistically marginalized in expectation, leaving scaffold-level shared variation as the dominant signal across clusters.
Since scaffold identity represents a stable environmental factor shared by all molecules in a cluster, its effect manifests as systematic shifts in the cluster-level mean label and is therefore captured by the context branch.
In contrast, the causal branch encodes sample-specific residual variation and becomes approximately invariant after within-cluster averaging, resulting in its diminished correlation with the aggregated labels.

Notably, for certain scaffold-dominated properties such as solubility, the causal structure can be inherently multi-scale.
Beyond functional-group-level polarity, molecule-specific structural factors related to scaffold geometry—such as steric effects or substitution patterns—may introduce residual variation within scaffold clusters.
The causal branch may retain weak but non-negligible correlation in these cases, reflecting remaining sample-intrinsic effects in practice rather than leakage of scaffold-level shared information.

These results indicate that, under scaffold-level marginalization, CLaP correctly allocates environment-level structural effects to the context branch while preserving molecule-specific variation in the causal branch in this analysis as intended by design.

% \begin{figure}[h]
%   \centering
%   \includegraphics[width=1.0\linewidth]{kdd2026/figures/causal_analysis_across_scaffolds.png}

%   \caption{causal_analysis_across_scaffolds.}
%   \label{fig:causal_analysis_across_scaffolds}
% \end{figure}

% \begin{figure}[h]
%   \centering
%   \includegraphics[width=1.0\linewidth]{kdd2026/figures/rebatching_correlation_2.png}

%   \caption{Environment interventions via re-batching.}
%   \label{fig:test-rebatch-global}
% \end{figure}

\subsubsection{Causal Invariance under Re-batching} 
\label{sec:test-rebatch-sensitivity}

\paragraph{Evaluation setup.}
To test whether the learned causal representation exhibits invariance under interventions on the environment, we conduct a controlled evaluation by manipulating the test-time batching scheme.
The batch construction is treated explicitly as an environment variable $E$. Concretely, we intervene on $E$ by altering how the same test examples are grouped into mini-batches,
with $E=(B,\text{shuffle})$, where $B\!\in\!\{8,16,32\}$ and shuffling is applied or not.
This realizes a family of interventions $\mathsf{do}(E=e)$ that perturb batch boundaries and co-occurrence statistics,
while preserving the underlying sample-intrinsic signal $c(x)$ by construction.

Under the batch-wise invariance formulation in Sec.~\ref{sec:batch-invariance},
such interventions affect only the batch-dependent context component $q(x,E)$ in the local predictor
$s(x,E)=ac(x)+bq(x,E)$ (Eq.~\eqref{eq:bi-pred}), while leaving $c(x)$ unchanged.
The depth-wise correlation schedule in our training objective
(Eqs.~\eqref{eq:corr}--\eqref{eq:Lcorr}) is designed to drive $b \rightarrow 0$,
encouraging the causal branch to depend on $c(x)$ and remain approximately invariant under $\mathsf{do}(E=e)$.

\paragraph{Observed behavior.}
For each intervention, we compute the \emph{global} Pearson correlation between the causal prediction $y_{\mathrm{c}}^\star$
and the ground-truth labels over the entire test set.
As shown in Fig.~\ref{fig:causal_rebaching}, the resulting correlation--depth curves remain tightly aligned
across all batch sizes and shuffling configurations in practice.

At the final layer, correlations concentrate within a narrow range,
indicating that the causal branch is insensitive to how test-time environments are constructed.
Minor deviations from the anchor correlation can be attributed to context-induced effects,
such as scaffold composition or other batch-level statistics.
Importantly, these effects are explicitly modeled and absorbed by the context branch,
allowing the causal path to preserve an invariant predictive signal across interventions. Since the interventions modify the environment variable $E$ while keeping the underlying examples fixed,
the observed stability provides direct evidence that the learned causal representation satisfies
the desired invariance property.

Together with the structural environment analysis, these two experiments provide complementary evidence of causal behavior. The structural analysis verifies that environment effects are allocated to the context branch, while the re-batching intervention demonstrates that the causal branch remains invariant under environment shifts, confirming causal–context disentanglement in CLaP.

% This pattern is consistent with our design in Sec.~\ref{sec:training}. The model benefits from \emph{complementarity} across modalities: the split routes label-relevant signal to the causal branch while the fusion gate selects the modality that carries the most batch-invariant evidence at each depth. Forcing embeddings to be too similar collapses modality-specific subspaces, reduces the gate’s effective choice set, and makes batch-coupled shortcuts more likely to be shared. The correlation curriculum then has less diverse causal evidence to promote, and the residual path absorbs less of the confounding context. In practice, we keep $\lambda_{\text{cons}}$ at zero or very small to preserve complementarity. 

\begin{figure*}[h]
  \centering
  \includegraphics[width=1.04\textwidth]{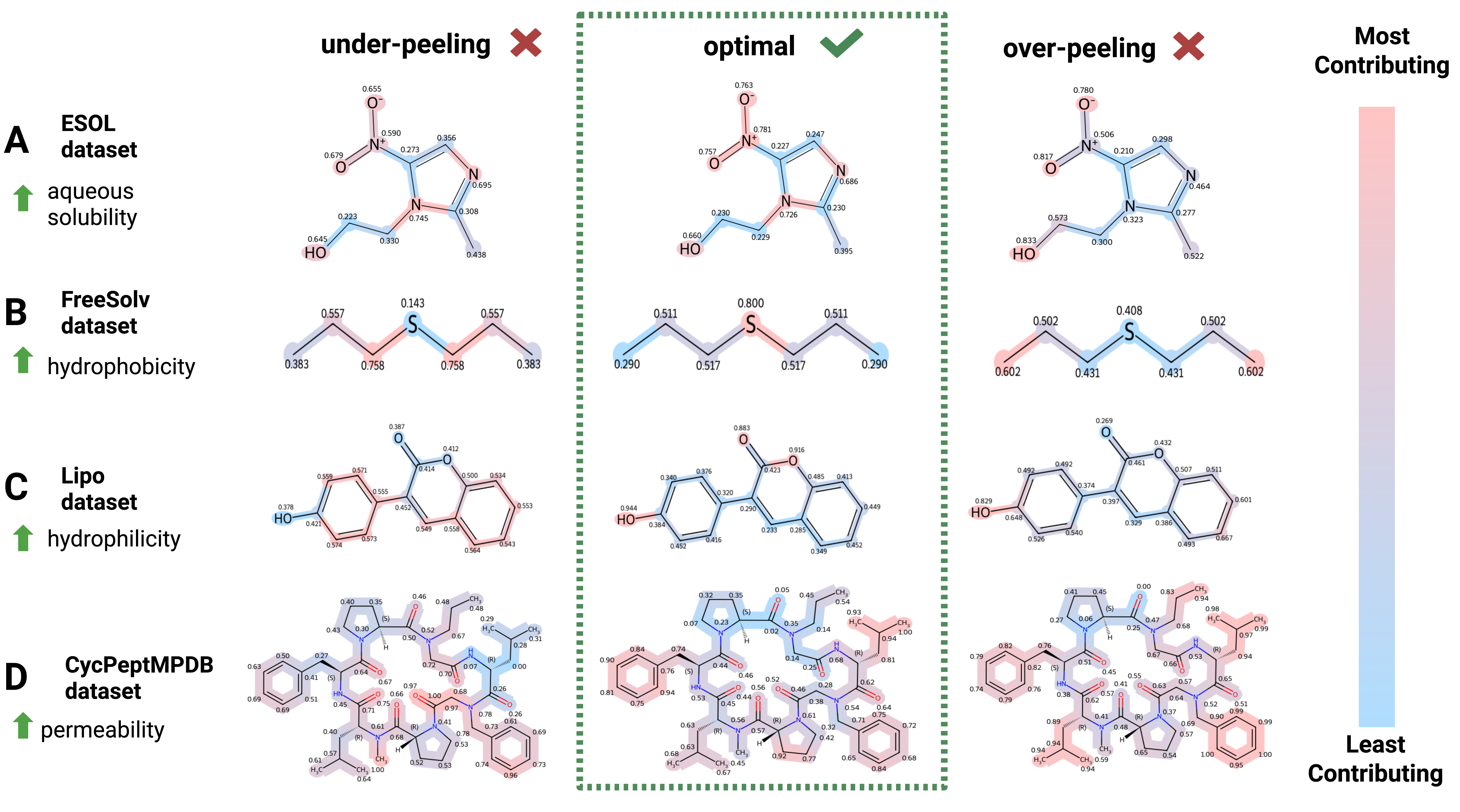}
  \caption{Layer-wise attribution maps in a multimodal setting across four benchmark datasets. Each row corresponds to a dataset and target property (green arrow). Three regimes are shown: under-peeling (left), optimal (center), and over-peeling (right). Atom-level contributions are visualized, with red indicating positive attribution to the target property and blue indicating lower contribution. Optimal peeling yields clearer and more interpretable attribution than under- or over-peeling.}
  \label{fig:case-study}
\end{figure*}

\subsection{Case study}

\subsubsection{Layer-wise Causal Attribution}

The causal split mechanism generates causal weights for each input component. Each causal block applies node-wise gating on atoms, producing
$\alpha_{i}^{(\ell,m)}\!\in\![0,1]$ for atom/group $i$ in modality
$m\!\in\!\{\textit{SM},\textit{PE},\textit{GE}\}$ at depth $\ell$.
We take the final-layer weight as the causal score,
$\pi_{i}^{(m)} = \alpha_{i}^{(L,m)} \in [0,1]$, and render \emph{causal saliency maps} $\{\pi_{i}^{(m)}\}$ to inspect behavior.

Higher $\pi_{i}^{(m)}$ indicates atoms routed through the causal branch at the final layer.
Figure~\ref{fig:case-study} compares under-peeling, optimal-peeling, and over-peeling regimes.
In the optimal regime, saliency concentrates on chemically meaningful moieties, consistent with chemical intuition. Across datasets, the resulting saliency patterns align with known structure--property relationships.
\textbf{A. ESOL} emphasizes polar nitro groups and heteroatoms, reflecting their role in hydrogen bonding and aqueous solubility.
\textbf{B. FreeSolv} assigns higher weight to hydrocarbon segments and the thio-linker, consistent with increased hydrophobicity.
\textbf{C. Lipo} highlights phenolic OH substituents and ring heteroatoms while down-weighting extended aromatic surfaces, capturing the balance between polar functionalities and nonpolar bulk.
\textbf{D. CycPeptMPDB} attributes importance to contiguous aromatic and aliphatic side chains that promote passive permeation, with reduced emphasis on backbone hydrogen-bond donors and acceptors.
Overall, these trends illustrate the intended peeling behavior. Shallow layers under-localize attribution, deep layers over-peel and diffuse it, while intermediate layers isolate the chemically relevant substructures that drive each property.

\subsubsection{Counterfactual Study}
\begin{figure}[h]            
  \centering
  \includegraphics[width=1\linewidth]{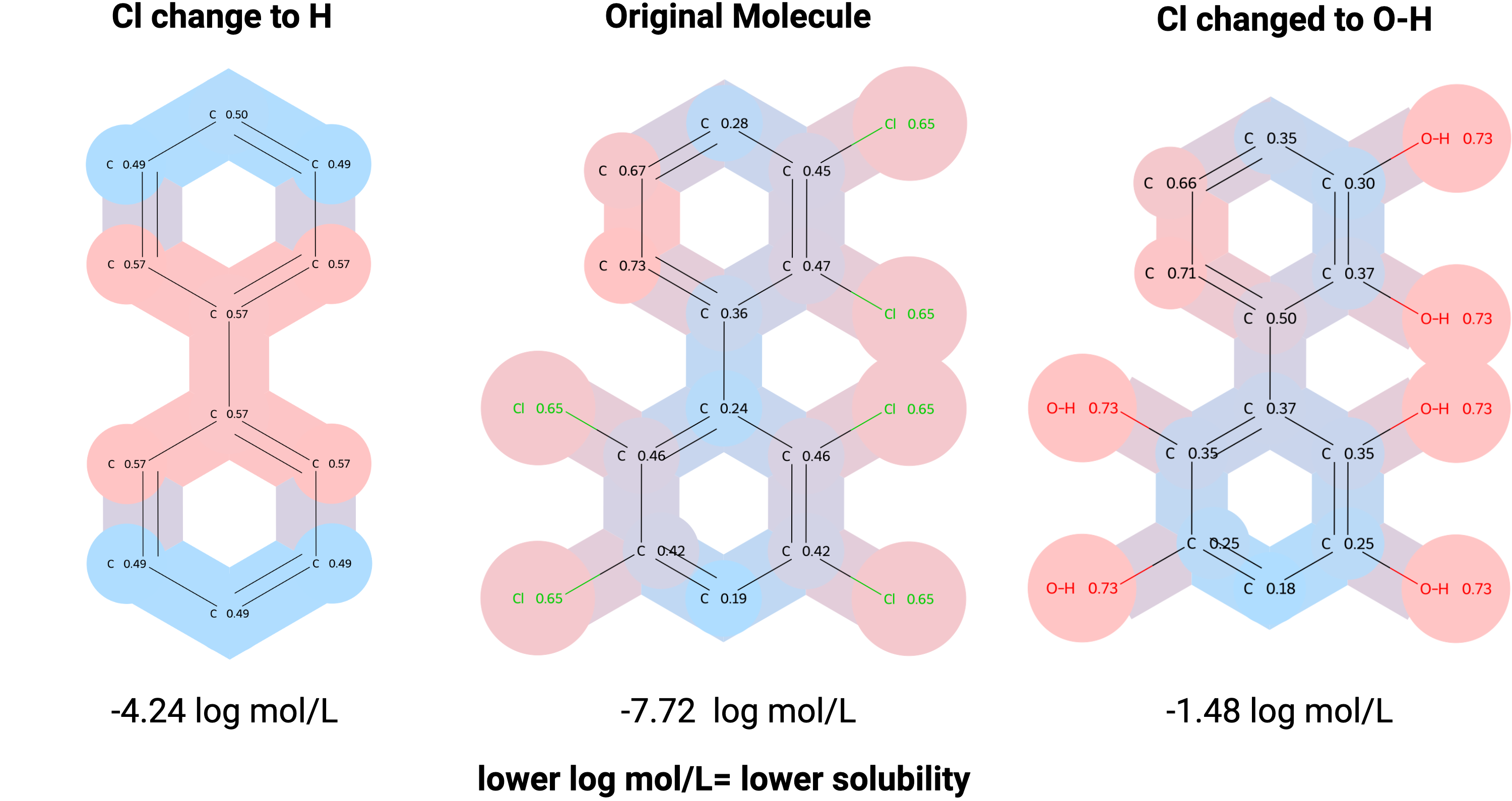}
\caption{Atom-level counterfactual explanation and attribution for ESOL solubility prediction produced by CLaP.}

  \label{fig:counterfactual_case-study}
\end{figure}

To further validate the learned causal weights, we perform a counterfactual experiment by substituting the highest-weight atoms with more soluble functional groups (Figure~\ref{fig:counterfactual_case-study}), and examine both solubility shifts and causal-weight responses. In the original molecule, the \texttt{Cl} atoms receive the highest causal scores, indicating that the model identifies them as solubility determinants. Replacing \texttt{Cl} with \texttt{H} removes these high-impact sites and leads to an increase in solubility. In contrast, substituting \texttt{Cl} with an \texttt{O--H} group preserves the same causal positions but reverses their chemical effect, producing an even larger increase. These counterfactual edits demonstrate that the model not only locates scaffold-level “causal sites,” but also correctly predicts both the direction and magnitude of property changes induced by specific chemical substitutions, reflecting counterfactual reasoning \cite{wang2024rlhex}.

\section{Related Work}

\subsection{Graph causal inference.}
In graph causal inference, recent work aims to improve robustness and interpretability in graph learning by separating label-relevant causal structure from spurious contextual signals \cite{an2025csib, yao24o, li24s}. DIR selects invariant rationales—subgraphs that remain predictive across environments—and then predicts with the remainder~\citep{DIR}. CIGA instead samples subgraphs via a learnable generator and enforces risk invariance through an information bottleneck~\citep{CIGA}.

In molecular settings, environment construction is simulated by adding or substituting functional groups and treating these variants as separate environments~\citep{li2025soft}. However, manipulations may be chemically unrealistic and exacerbate the distribution gap between training and testing data. CAL alleviates shortcut learning and improves interpretability~\citep{CAL}, yet prior work focuses on classification. Regression is inherently more challenging: the model must precisely localize causal substructures rather than rely on class-level artifacts or thresholds.
Most recently, CGL extends causal graph learning to regression by jointly modeling causal and confounding subgraphs, with a particular focus on molecular property prediction~\citep{CGL}.

However, instead of crafting synthetic environments, our proposed CLaP achieves batch-wise invariance via the natural mini-batch reshuffling during training, yielding stable within-batch correlations across depths and better causal capture under re-batching.

\subsection{Molecular property prediction.}
Neural architectures have become a cornerstone for molecular property prediction~\citep{Gilmer2017Neural,Wu2018MoleculeNet}, achieving performance by learning molecular representations. Beyond classical message passing, recent architectures incorporate complex structural priors: transformer-based models such as MAT encode geometric biases via inter-atomic distances~\citep{MAT}, while multimodal frameworks fuse chemical language with graph structure for richer cross-domain context~\citep{Rollins2024MolPROP, Vinh2025SMTP, liu2024git_mol, luo2025molcl_sp}.

Despite these advances, most models behave as black boxes, capturing correlations without disentangling causal factors, which limits interpretability and usefulness in rational molecular design~\citep{JimenezLuna2020XAI, Vinh2025BeyondPrediction}. To address this, several recent works enhance representation quality by incorporating structural insights or chemical priors. GSL-MPP leverages relational structure across molecules to improve predictive performance~\citep{GSL}, while MolFCL introduces fragment-aware contrastive learning and functional-group-level priors to obtain chemically robust and interpretable representations~\citep{MolFCL}.

\section{Discussion}

We introduced CLaP, a layerwise causal–context peeling framework that enforces batch-wise invariance to separate label-relevant structure from contextual shortcuts. This design improves predictive performance while yielding consistent atom-level explanations.

Beyond the current setting, the principle of depthwise causal alignment is broadly applicable. The framework can extend to classification via within-class correlation objectives and integrate into transformer architectures to route tokens into causal and context pathways, offering a general mechanism for interpretable and robust representation learning across structured prediction tasks.

% \section*{Reproducibility statement}
% \noindent
% Implementation details and hyperparameters are in Appendix~\ref{sec:implementation_details}. The supplementary material includes a code archive with configs, preprocessing, data loaders, architectures, and evaluations.

% \section*{Ethics statement}
% \noindent
% We adhere to KDD Code of Ethics. No human subjects or personally identifiable data are involved. To reduce misuse risk, we release benchmarking code and models only for these datasets and provide no synthesis or activity oracles. No conflicts of interest or sensitive sponsorships are present.

\newpage
\bibliographystyle{ACM-Reference-Format}
\bibliography{biblio}
\appendix
\newpage

\section*{Appendix Contents}

\noindent\textbf{1. Theoretical Analysis} \dotfill \pageref{app:bi-analysis} \\
\hspace*{1em} 1.1 Batch-Wise Invariance Analysis \dotfill \pageref{app:bi-analysis} \\
\hspace*{1em} 1.2 Context Branch Analysis \dotfill \pageref{app:Context_Branch_analysis} \\
\hspace*{1em} 1.3 Target-Correlation Analysis \dotfill \pageref{app:target-pressure}

\vspace{0.5em}

\noindent\textbf{2. Additional Experiments} \dotfill \pageref{app:Additional-Experiments} \\
\hspace*{1em} 2.1 Effect of Peeling Hyperparameters \dotfill \pageref{app:Peeling-Hyperparameters} \\
\hspace*{1em} 2.2 Additional Ablation Study \dotfill \pageref{app:Additional-Ablation}

\vspace{0.5em}

\noindent\textbf{3. Additional Case Studies and Multimodality Analysis} \dotfill \pageref{fig:esolap1} \\
\hspace*{1em} 3.1 ESOL Dataset \dotfill \pageref{fig:esolap1} \\
\hspace*{1em} 3.2 FreeSolv Dataset \dotfill \pageref{fig:freesolvap1} \\
\hspace*{1em} 3.3 Lipo Dataset \dotfill \pageref{fig:lipoap1} \\
\hspace*{1em} 3.4 CycPeptMPDB Dataset \dotfill \pageref{fig:pep1}

\vspace{0.5em}

\noindent\textbf{4. Implementation Details} \dotfill \pageref{sec:implementation_details} \\
\hspace*{1em} 4.1 Proposed Framework Implementation Details \dotfill \pageref{sec:implementation_details} \\
\hspace*{1em} 4.2 Baselines Implementation Details \dotfill \pageref{sec:baseline_implementation_details} \\

\newpage

\section*{Appendix}

\section{Theoretical Analysis}

\subsection{Batch-Wise Invariance Analysis}
\label{app:bi-analysis}

We use the batch-centered correlation in Eq.~\eqref{eq:bi-inv} and the local form
\(s(x,e)=a\,c(x)+b\,q(x,e)\) in Eq.~\eqref{eq:bi-pred}.
Define within-batch moments
\[
\begin{gathered}
\sigma_{c,e}^{2}=\mathrm{Var}_e(c),\quad
\sigma_{q,e}^{2}=\mathrm{Var}_e(q),\quad
\sigma_{y,e}^{2}=\mathrm{Var}_e(y),\\
\kappa=\mathrm{Corr}_e(c,y),\quad
\alpha_e=\mathrm{Corr}_e(q,y),\quad
\rho_e=\mathrm{Corr}_e(c,q)
\end{gathered}
\]
with \(\kappa\) constant in \(e\) by (A1). Set
\[
A_e = a\,\sigma_{c,e}, \qquad
B_e = b\,\sigma_{q,e}.
\]

\noindent\textbf{Closed form for \(\mathrm{Corr}_e(s,y)\).}
Starting from the definition,
\begin{align}
\mathrm{Corr}_e(s,y)
&= \frac{\mathrm{Cov}_e(s,y)}{\sqrt{\mathrm{Var}_e(s)\,\mathrm{Var}_e(y)}}
\label{eq:app-corr-start}\\
&= \frac{\mathrm{Cov}_e(a c + b q,\,y)}{\sqrt{\mathrm{Var}_e(a c + b q)\,\mathrm{Var}_e(y)}} \nonumber\\
&= \frac{a\,\mathrm{Cov}_e(c,y) + b\,\mathrm{Cov}_e(q,y)}
         {\sqrt{\,\mathrm{Var}_e(a c + b q)\,}\;\sqrt{\mathrm{Var}_e(y)}} \nonumber\\
&= \frac{a\,\sigma_{c,e}\,\sigma_{y,e}\,\kappa \;+\; b\,\sigma_{q,e}\,\sigma_{y,e}\,\alpha_e}
         {\sqrt{\,a^{2}\sigma_{c,e}^{2} + b^{2}\sigma_{q,e}^{2} + 2ab\,\sigma_{c,e}\sigma_{q,e}\,\rho_e\,}\;\sigma_{y,e}}
\\
&= \frac{A_e\,\kappa + B_e\,\alpha_e}
         {\sqrt{A_e^{2}+B_e^{2}+2A_e B_e \rho_e}}
   \eqqcolon \frac{A_e\kappa + B_e\alpha_e}{D_e}.
\label{eq:app-corr}
\end{align}
Here \(D_e=\sqrt{A_e^{2}+B_e^{2}+2A_eB_e\rho_e}=\sigma_{s,e}>0\) is the within-batch standard deviation of \(s\).

\noindent\textbf{A monotonicity observation (\(\partial_{\alpha_e}\mathrm{Corr}_e\)).}
For fixed \(e\), \(A_e,B_e,\rho_e\) do not depend on \(\alpha_e\), hence from \eqref{eq:app-corr}
\begin{align}
\frac{\partial}{\partial \alpha_e}\,\mathrm{Corr}_e(s,y)
= \frac{\partial}{\partial \alpha_e}\,\frac{A_e\kappa + B_e\alpha_e}{D_e}
= \frac{B_e}{D_e}. \label{eq:mono-deriv}
\end{align}
Therefore if \(B_e\neq 0\) (equivalently \(b\neq 0\)), \(\mathrm{Corr}_e(s,y)\) is strictly monotone in \(\alpha_e\).

\noindent\textbf{Consequence under exact invariance (\(\mathrm{Corr}_{e_1}=\mathrm{Corr}_{e_2}\)).}
Assume there exist batches \(e_1,e_2\) with
\(\sigma_{c,e_1}=\sigma_{c,e_2}\),
\(\sigma_{q,e_1}=\sigma_{q,e_2}\),
\(\rho_{e_1}=\rho_{e_2}\),
but \(\alpha_{e_1}\neq \alpha_{e_2}\).
Then \(A_{e_1}=A_{e_2}\), \(B_{e_1}=B_{e_2}\), \(D_{e_1}=D_{e_2}\), and
\[
\mathrm{Corr}_{e_1}(s,y)-\mathrm{Corr}_{e_2}(s,y)
=\frac{B_{e_1}}{D_{e_1}}\big(\alpha_{e_1}-\alpha_{e_2}\big).
\]
If \(\mathrm{Corr}_e(s,y)\) is identical across batches (zero dispersion in Eq.~\eqref{eq:bi-inv}), the right-hand side must be \(0\).
Since \(\alpha_{e_1}\neq\alpha_{e_2}\), this forces \(B_{e_1}=0\), hence \(b=0\) because \(\sigma_{q,e_1}>0\) by (A3).

\noindent\textbf{Approximate invariance via a first-order expansion.}
Let \(\delta_e(b):=\mathrm{Corr}_e(s,y)-\rho\), where \(\rho\) is the target in Eq.~\eqref{eq:bi-inv}.
Write \(\mathrm{Corr}_e(s,y)=\Phi_e(B_e)\) with
\(\Phi_e(B)=\dfrac{A_e\kappa + B\alpha_e}{\sqrt{A_e^{2}+B^{2}+2A_e B \rho_e}}\).
By the quotient rule,
\[
\Phi_e'(B)
= \frac{\alpha_e\,D_e - (A_e\kappa + B\alpha_e)\,D_e'}{D_e^{2}},
\qquad
D_e' = \frac{B + A_e\rho_e}{D_e}.
\]
Evaluating at \(B=0\) (so \(D_e=|A_e|\)) gives
\begin{align}
\left.\frac{\partial}{\partial B}\,\mathrm{Corr}_e(s,y)\right|_{B=0}
= \Phi_e'(0)
= \frac{\alpha_e A_e^{2} - A_e\kappa(A_e\rho_e)}{|A_e|^{3}}
= \frac{\alpha_e - \kappa\rho_e}{|A_e|}. \label{eq:dCorr-dB}
\end{align}
Since \(B_e=b\,\sigma_{q,e}\), the chain rule yields
\begin{equation}
\left.\frac{\partial}{\partial b}\,\mathrm{Corr}_e(s,y)\right|_{b=0}
= \sigma_{q,e}\,\left.\frac{\partial}{\partial B}\,\mathrm{Corr}_e(s,y)\right|_{B=0}
= \frac{\sigma_{q,e}}{|A_e|}\,\big(\alpha_e-\kappa\rho_e\big).
\label{eq:dCorr-db}
\end{equation}
Therefore, for small \(b\),
\begin{equation}
\delta_e(b)
= \delta_e(0) + b\,\Gamma_e + o(b),
\qquad
\Gamma_e = \frac{\sigma_{q,e}}{|A_e|}\,\big(\alpha_e-\kappa\rho_e\big).
\label{eq:linearization}
\end{equation}
If \(\sum_e \Gamma_e^2>0\) (which holds whenever \(\alpha_e\) varies across batches by (A2) and \(q\) is not batchwise collinear with \(c\)),
minimizing the dispersion surrogate \(\sum_e \delta_e(b)^2\) in Eq.~\eqref{eq:bi-sur} yields the closed-form
\begin{equation}
b^\star
= -\,\frac{\sum_e \delta_e(0)\,\Gamma_e}{\sum_e \Gamma_e^2}\;+\;o(1).
\label{eq:bstar}
\end{equation}

\noindent\textbf{Invariance conclusion.}
At \(b=0\), from \eqref{eq:app-corr} we have
\[
\mathrm{Corr}_e(s,y)\big|_{b=0}
=\frac{A_e\kappa}{|A_e|}\,,
\quad\Rightarrow\quad
\delta_e(0)=\frac{A_e\kappa}{|A_e|}-\rho.
\]
With the natural choice \(\rho=\kappa\) and \(a\ge 0\) (so \(A_e\ge 0\)), we get \(\delta_e(0)=0\) for all \(e\).
Moreover, from \eqref{eq:linearization} at \(b^\star\) we have
\(\delta_e(b^\star)=b^\star \Gamma_e + r_e(b^\star)\) with \(\sum_e r_e(b^\star)^2=o\!\left((b^\star)^2\right)\).
Taking the $\ell_2$-norm and using Cauchy--Schwarz,
\[
|b^\star|\,\|\Gamma\|_2 \;\le\; \|\delta(b^\star)\|_2 + \|r(b^\star)\|_2
\;=\; o\!\left(|b^\star|\right),
\]
and since \(\|\Gamma\|_2>0\) by (A2)–(A3), it follows that \(b^\star \Rightarrow\ 0\) as the model gradually reduces the contribution of the context component. Hence in the limit the predictor discards the batch-dependent \(q\) (i.e., \(b^\star\Rightarrow\ 0\) so \(s \approx\ a\,c\)), so
\[
\mathrm{Corr}_e(s,y)\ \approx \
\mathrm{Corr}_e(c,y)=\kappa
\qquad \text{for all } e.
\]

\subsection{Context Branch Analysis}
\label{app:Context_Branch_analysis}

With squared loss and the stop–gradient in Eq.~\eqref{eq:Lctx}, training $t$ treats $y_{\mathrm{c}}^\star$ as a constant with respect to its parameters, so the context head solves the conditional least-squares problem
\[
t^\star(x)
=\arg\min_t \ \E\!\big[(y-y_{\mathrm{c}}^\star-t(x))^2\mid x\big]
=\E\!\big[y-y_{\mathrm{c}}^\star\mid x\big].
\]
That is, the \emph{best} residual correction at input $x$ is the \emph{conditional mean} of the residual.

\medskip
Let $r:=y-y_{\mathrm{c}}^\star$. By the definition of MSE risk $\mathcal{R}(f)=\E[(y-f(x))^2]$,
\[
\mathcal{R}(y_{\mathrm{c}}^\star)=\E[r^2].
\]

Applying the Law of total expectation and conditional variance decomposition with $t\equiv 0$ gives the decomposition of the risk of $y_{\mathrm{c}}^\star$:
\[
\mathcal{R}(y_{\mathrm{c}}^\star)=\E[r^2]
=\E\!\big[\Var(r\mid x)\big]+\E\!\big[(\E[r\mid x])^2\big],
\]
which splits the error into (i) irreducible noise $\E[\Var(r\mid x)]$ and (ii) predictable bias $\E[(\E[r\mid x])^2]$.

Plugging $t^\star(x)=\E[r\mid x]$ into the predictor $\widehat y=y_{\mathrm{c}}^\star+t$ removes the predictable bias:
\[
\mathcal{R}(y_{\mathrm{c}}^\star+t^\star)=\E\!\Big[\Var(r\mid x)+(\E[r\mid x]-t^\star(x))^2\Big]\overset{t^\star=\E[r\mid x]}{=}\E\!\big[\Var(r\mid x)\big].
\]
So adding $t^\star$ strictly reduces risk by $\E[(\E[r\mid x])^2]\!\ge\!0$ whenever the residual is partially predictable from $x$.

\subsection{Target-Correlation Analysis}
\label{app:target-pressure}

Assume the additive label model
\begin{equation}
y=\theta\,c(x)+\eta,\qquad \E[\eta\mid x]=0,\quad \Var(\eta)>0, \quad  \theta \ge 0
\label{eq:tcl-model}
\end{equation}
Then
\begin{align}
\mathrm{Corr}(c,y)
= \frac{\Cov(c,y)}{\sqrt{\Var(c)\Var(y)}}
= \frac{\theta\,\Var(c)}{\sqrt{\Var(c)\,(\theta^2\Var(c)+\Var(\eta))}}
=: \kappa \;<\; 1.
\label{eq:tcl-kappa}
\end{align}
Any batch-invariant predictor using only \(c\) has \(s(x,e)=a\,c(x)\) and thus \(\mathrm{Corr}_e(s,y)=\kappa\) for all \(e\).

Let \(s(x,e)=a\,c(x)+b\,q(x,e)\) as in Eq.~\eqref{eq:bi-pred}, and define
\[
\begin{aligned}
\sigma_{c,e}^2 &= \Var_e(c), \quad \sigma_{q,e}^2 = \Var_e(q), \quad
\kappa = \Corr_e(c,y), \\
\alpha_e &= \Corr_e(q,y), \quad \rho_e = \Corr_e(c,q), \quad
A_e = a\,\sigma_{c,e}
\end{aligned}
\]
From Eq.~\eqref{eq:app-corr},
\[
\Corr_e(s,y)=\frac{A_e\,\kappa + b\,\sigma_{q,e}\,\alpha_e}
                  {\sqrt{A_e^2 + b^2\sigma_{q,e}^2 + 2A_e b \sigma_{q,e}\rho_e}}.
\]
For the dispersion surrogate \(\mathcal{L}_\rho(a,b)=\sum_e\big(\mathrm{Corr}_e(s,y)-\rho_\star\big)^2\) with target \(\rho_\star\in(0,1)\),
the first-order sensitivity at the invariant point \(b=0\) is (Appendix~\ref{app:bi-analysis})
\begin{equation}
\begin{aligned}
\left.\frac{\partial}{\partial b}\Corr_e(s,y)\right|_{b=0}
&= \frac{\sigma_{q,e}}{|A_e|}\,\big(\alpha_e-\kappa\,\rho_e\big)
\ \eqqcolon\ \Gamma_e, \\
\left.\frac{\mathrm{d}}{\mathrm{d} b}\mathcal{L}_\rho\right|_{b=0}
&= 2(\kappa-\rho_\star)\sum_e \Gamma_e .
\end{aligned}
\label{eq:tcl-grad}
\end{equation}
Under (A1)–(A3) and the mild non-collinearity \(\exists e:\alpha_e\neq \kappa\rho_e\),
the sum \(\sum_e\Gamma_e\) is generically nonzero. If \(\rho_\star>\kappa\) then \(\kappa-\rho_\star<0\), so
\(\mathrm{d}\mathcal{L}_\rho/\mathrm{d}b|_{b=0}\neq 0\) and \(b=0\) is not a local minimizer.
Hence any optimizer of \(\mathcal{L}_\rho\) satisfies \(|b|>0\): the predictor must recruit the batch-dependent \(q(x,e)\)
to reduce the surrogate when the target exceeds the invariant limit \(\kappa\).

Moreover, writing \(\delta_e(b)=\mathrm{Corr}_e(s,y)-\rho_\star\) and expanding at \(b=0\) gives
\[
\delta_e(b)=(\kappa-\rho_\star)+b\,\Gamma_e+o(b),
\quad\Rightarrow\quad
b^\star
= \frac{(\rho_\star-\kappa)\,\sum_e \Gamma_e}{\sum_e \Gamma_e^2}\;+\;o(1).
\]
Thus the leakage magnitude grows linearly with the target gap \((\rho_\star-\kappa)\) and is controlled by the batch statistics through \(\{\Gamma_e\}\).

\section{Additional Experiments}
\label{app:Additional-Experiments}

\subsection{Effect of Peeling Hyperparameters}
\label{app:Peeling-Hyperparameters}

We analyze how the peeling configuration, parameterized by the depth $L$ and the target correlation $\rho_{\max}$, influences performance.
Both $L$ and $\rho_{\max}$ are treated as tunable hyperparameters, and we observe that extreme settings can induce \emph{under}- or \emph{over}-peeling.
Empirically, we find that the optimal configurations of $L$ and $\rho_{\max}$ typically occur at relatively shallow depths, indicating that effective settings can be identified with limited hyperparameter tuning effort.

\noindent\textbf{Peeling depth.}
We first vary the number of causal blocks \(L \in \{3,5,7,9\}\) and observe a U-shaped response in MAE/MSE on CycPeptMPDB in the left panel of Table~\ref{tab:peeling_design}. With too few blocks ($L{=}3$), the hierarchy \emph{under-peels}. It lacks capacity to progressively strip context, leaving causal and context signals entangled and hurting accuracy. With too many blocks ($L{=}7,9$), the hierarchy \emph{over-peels}. The longer optimization path increases variance and can misroute label-relevant patterns into the context stream, degrading generalization. An interior optimum at $L{=}5$ aligns with a bias–variance view: depth must be large enough to separate signals but not so large that true causal evidence is shaved off together with context.

\begin{table}[h]
\centering
\caption{Layers and $\rho_{\max}$ vs.\ performance on CycPeptMPDB.}
\label{tab:peeling_design}

\setlength{\tabcolsep}{4pt}
\renewcommand{\arraystretch}{0.8}

\begin{tabularx}{\columnwidth}{lCCCCC}
\toprule
\multicolumn{3}{c}{\textit{\textbf{Peeling Depth ($L$)}}} &
\multicolumn{3}{c}{\textit{\textbf{Peeling Strength ($\rho_{\max}$)}}} \\
\cmidrule(lr){1-3}\cmidrule(lr){4-6}
$L$ & MAE & MSE & $\rho_{\max}$ & MAE & MSE \\
\midrule
3          & 0.3265 & 0.1903 & 0.60        & 0.3131 & 0.1741 \\
\rowcolor{gray!10}
\textbf{5} & \textbf{0.3056} & \textbf{0.1644} & 0.70        & 0.3162 & 0.1733 \\
7          & 0.3296 & 0.2020 & \textbf{0.80} & \textbf{0.3056} & \textbf{0.1644} \\
\rowcolor{gray!10}
9          & 0.3297 & 0.1875 & 0.90        & 0.3133 & 0.1834 \\
\bottomrule
\end{tabularx}
\end{table}

\noindent\textbf{Peeling strength.}
We examine the effect of the target correlation parameter $\rho_{\max}\!\in\!\{0.60,0.70,0.80,0.90\}$ on CycPeptMPDB in the right panel of Table~\ref{tab:peeling_design}. Smaller targets (0.60–0.70) \emph{under-peel}. The causal branch is not pushed hard enough, so contextual residue persists, and errors rise.
A moderate target (0.80) yields the best trade-off. Pursuing near-perfect within-batch correlation (0.90) \emph{over-peels}. To satisfy an aggressive target, the model exploits batch-coupled features that should remain context, increasing variance and hurting generalization. Appendix~\ref{app:target-pressure} provides additional evidence that an overly large target correlation causes context leakage.

The proposed causal split mechanism generates explicit causal weights for each input component. Each causal block applies node-wise gating on atoms, producing
$\alpha_{i}^{(\ell,m)}\!\in\![0,1]$ for atom/group $i$ in modality
$m\!\in\!\{\textit{SM},\textit{PE},\textit{GE}\}$ at depth $\ell$.
We take the final-layer weight as the causal score,
$\pi_{i}^{(m)} = \alpha_{i}^{(L,m)} \in [0,1]$, and render \emph{causal saliency maps} $\{\pi_{i}^{(m)}\}$ to inspect learned behavior. 

Higher $\pi_{i}^{(m)}$ indicates atoms routed through the causal branch at the final layer. Figure~\ref{fig:case-study} compares the under-peeling, optimal-peeling, and over-peeling settings. In the optimal case, saliency concentrates on chemically meaningful moieties, \textit{consistent with chemical intuition}. At the optimal depth, CLaP concentrates attribution on chemically coherent drivers per task.
Across the four datasets, \textbf{A. ESOL} highlights that polar nitro groups and heteroatoms consistently dominate, reflecting their strong role in hydrogen bonding and aqueous solubility (especially in water); \textbf{B. FreeSolv} shows that hydrocarbon segments together with the thio-linker carry the highest weights, aligning with sulfur’s tendency to raise local hydrophobicity and thus overall lipophilicity; and \textbf{C. Lipo} indicates that phenolic OH substituents and ring heteroatoms elevate hydrophilicity while the extended aromatic surface is down-weighted, capturing the trade-off between polar handles and nonpolar bulk consistent with established lipophilicity trends. Finally, in \textbf{D. CycPeptMPDB dataset}, contiguous aromatic and aliphatic side chains form lipophilic patches that promote passive permeation, whereas backbone hydrogen-bond donors and acceptors contribute less, consistent with the desolvation cost of membrane crossing. This pattern reflects the intended peel—the shallow layers under-localize the signal, the deep layers over-peel and smear attribution, and the middle layers isolate chemically relevant substructures that drive the property.

\subsection{Additional Ablation Study on Framework Design}
\label{app:Additional-Ablation}

\begin{table}[h]
\centering
\caption{Additional ablation study on framework design across four datasets. Best indicated in \textbf{Bold}.}
\label{tab:ablations_tail}

\renewcommand{\arraystretch}{1.0}

{
\resizebox{\columnwidth}{!}{%
\begin{tabular}{l*{4}{cc}}
\hline
\textbf{Setting} &
\multicolumn{2}{c}{\textit{ESOL}} &
\multicolumn{2}{c}{\textit{FreeSolv}} &
\multicolumn{2}{c}{\textit{Lipo}} &
\multicolumn{2}{c}{\textit{CycPeptMPDB}} \\
\cmidrule(lr){2-3}\cmidrule(lr){4-5}\cmidrule(lr){6-7}\cmidrule(lr){8-9}
 & MAE & MSE & MAE & MSE & MAE & MSE & MAE & MSE \\
\hline
Avg causal layer &
0.4798 & 0.4377 &
0.9612 & 1.5843 &
0.5621 & 0.5041 &
0.3227 & 0.1859 \\

\rowcolor{gray!10}
No mono penalty &
0.5692 & 0.5740 &
1.0241 & 1.7210 &
0.5288 & 0.4590 &
0.3137 & 0.1803 \\

\textbf{CLaP} &
\textbf{0.4456} & \textbf{0.3583} &
\textbf{0.7020} & \textbf{0.8866} &
\textbf{0.4672} & \textbf{0.3645} &
\textbf{0.3056} & \textbf{0.1644} \\
\hline
\end{tabular}%
}}
\end{table}

\noindent\textbf{w/o monotonicity constraint.}
The monotonicity penalty $\Ls_{\mathrm{mono}}$ (Eq.~\eqref{eq:Lmono}) complements the correlation loss by coupling adjacent depths, encouraging non-decreasing label–correlation as depth increases.
This depthwise ordering regularizes training, especially for deeper causal blocks, by discouraging oscillations and preventing later layers from overwriting gains established earlier.

Ablating \(\Ls_{\mathrm{mono}}\) weakens layerwise refinement of causal features, leading to a modest yet consistent performance drop. Monotonicity regularizer improves hierarchical stability and makes the layerwise specialization more reliable.

\noindent\textbf{Average causal across layers.}
This ablation probes the depthwise correlation design.
The targets $\{\rho_\ell\}$ increase with depth and, together with $\Ls_{\mathrm{corr}}$ and $\Ls_{\mathrm{mono}}$ (Eqs.~\eqref{eq:Lcorr}, \eqref{eq:Lmono}), train the deepest causal block to become progressively more label–relevant.
Here we replace the final-layer causal output with a uniform average of $\{\vc^{(\ell)}\}_{\ell=1}^{L}$. Given \(\mathrm{corr}_1\!\le\!\cdots\!\le\!\mathrm{corr}_L\) under \(\Ls_{\mathrm{corr}}+\Ls_{\mathrm{mono}}\), averaging \(\bar{c}=\tfrac{1}{L}\sum_\ell \vc^{(\ell)}\) mixes in low-correlation early representations.
For centered variables, \(\mathrm{Corr}(\bar{c},y)\le \mathrm{Corr}(\vc^{(L)},y)\), with strict inequality unless all \(\vc^{(\ell)}\) are colinear with \(\vc^{(L)}\).
Thus using \(\vc^{(L)}\) is theoretically favored.

In our experiments, averaging across layers undermines depthwise specialization and mixes early-layer representations that retain substantial context information into the final predictor, contaminating the causal signal and degrading performance (e.g. MSE: $0.164 \!\to\! 0.185$).
These results validate the design: the last causal layer concentrates the causal signal, while earlier layers serve as progressively coarser precursors.

Together, the correlation schedule shapes \emph{what} each depth should capture; the monotonicity term enforces \emph{how} the hierarchy evolves; the context branch determines \emph{where} non-causal residuals go.

\section{Additional Case Studies and Multimodality Analysis}

\subsection{ESOL Dataset}
We benchmark on the ESOL dataset, which contains \emph{1128 small organic molecules} with experimentally measured \emph{aqueous solubility} reported as
\[
\log S \equiv \log_{10}\!\big(\text{molar solubility in water (mol/L)}\big).
\]
Thus, the sign directly encodes scale: positive values mean solubility $>1$\,mol/L, $\log S{=}0$ corresponds to $1$\,mol/L, and negative values mean solubility $<1$\,mol/L. \\

\begin{figure}[h]
  \centering
  \includegraphics[width=\linewidth]{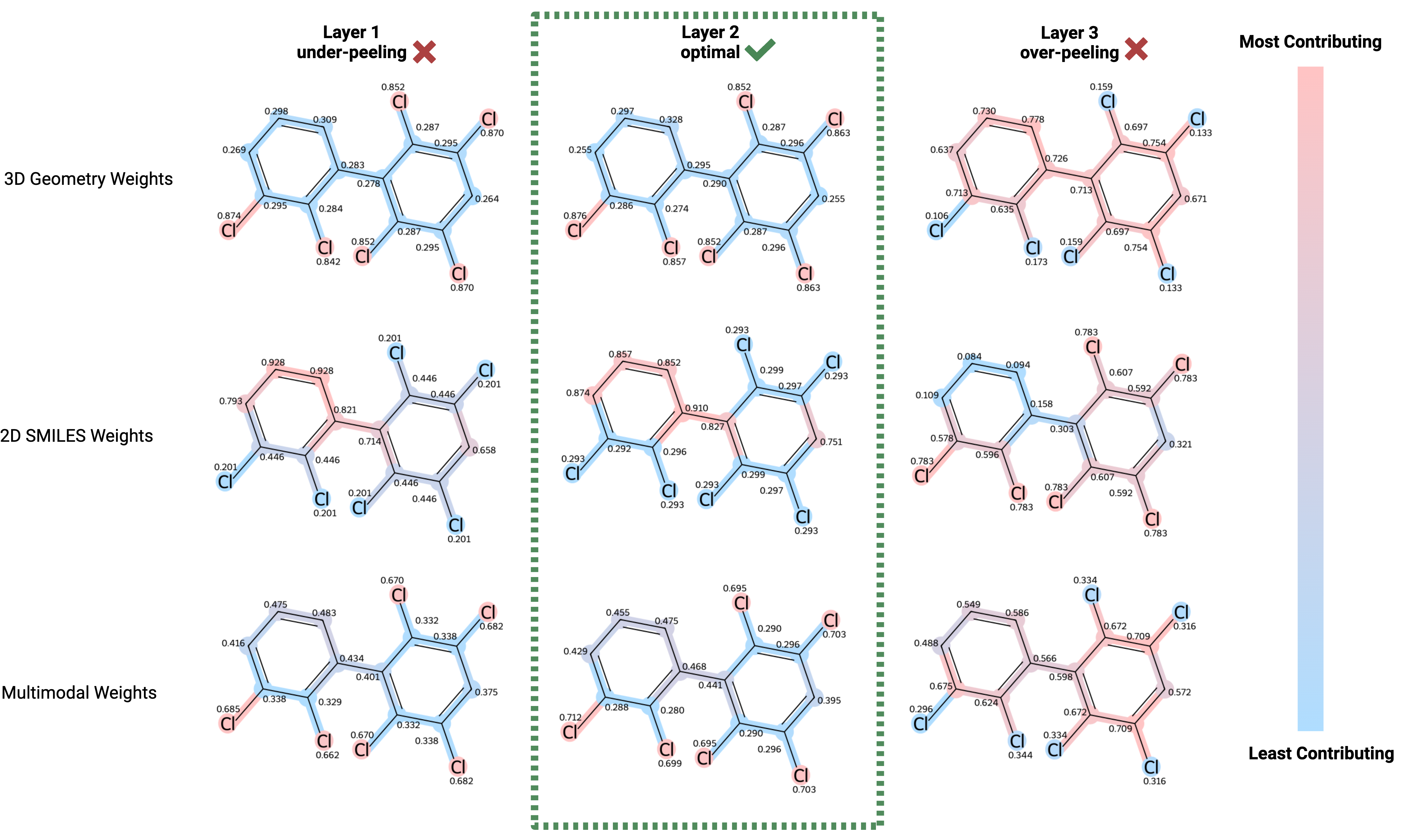}
  \caption{Layer-wise attribution maps across input modalities. Red colored atoms are attributed with affecting the molecule's water affinity the most.}
  \label{fig:esolap1}
\end{figure}

For Figure \ref{fig:esolap1}, at the optimal depth (L2), CLaP assigns the highest weights to chloro substituents and adjacent aryl carbons that cluster via short Cl–C/Cl–Cl distances in 3D; the fused readout preserves this through-space signal while the 2D branch highlights halogenated positions from connectivity alone. This depth-wise focus is exactly the intended “peel”: batch-coupled context is removed, and label-relevant structure (a single hydrophobic patch) is retained. The outcome matches physical chemistry: contiguous chlorination increases non-polar surface and reduces hydrogen-bonding capacity, driving logS lower. Under-peeling spreads attribution across the ring (shortcuts), and over-peeling reintroduces residual context, illustrating why CLaP’s layerwise schedule matters. 

\begin{figure}[h]
  \centering
  \includegraphics[width=\linewidth]{\detokenize{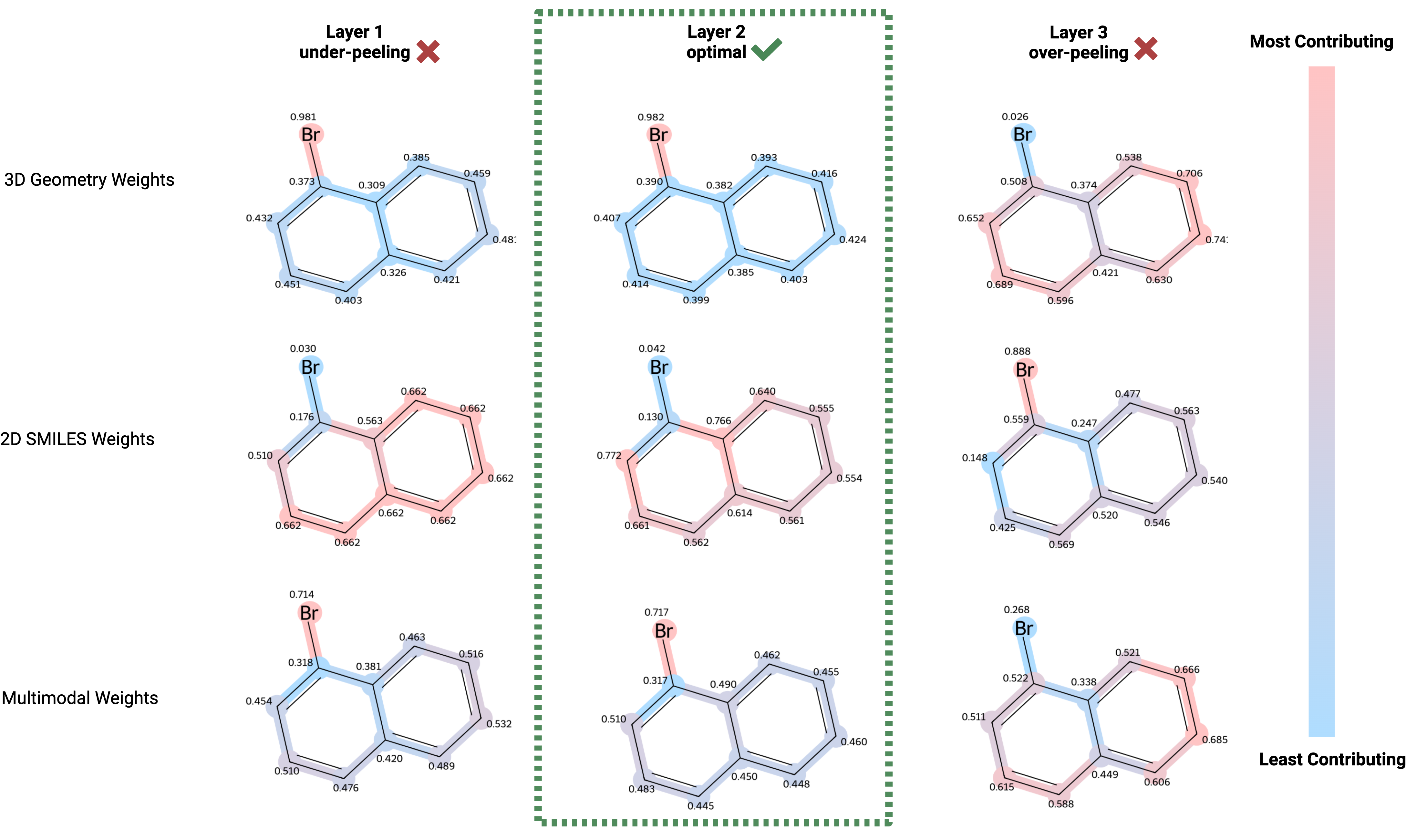}}
  \caption{Layer-wise attribution maps across input modalities. Red colored atoms are attributed with affecting the molecule's insolubility the most.}
  \label{fig:esolap2}
\end{figure}

For Figure \ref{fig:esolap2}, at the optimal depth (L2), CLaP assigns the highest weights to the \emph{aryl bromide} and the \emph{fused bicyclic ring junction}; the 3D branch aggregates the planar $\pi$-surface into a single hydrophobic patch, while the 2D branch captures bromination and ring fusion from connectivity alone. This depth-wise focus is exactly the intended ``peel'': batch-coupled context is removed, and label-relevant structure (a large, contiguous hydrophobic surface) is retained. The outcome matches physical chemistry: expanded aryl surface together with Br increases lipophilicity and contributes essentially no hydrogen-bond donors/acceptors, yielding minimal polar counterbalance and driving $\log S$ more negative (lower aqueous solubility). Under-peeling spreads attribution across the scaffold (shortcuts), and over-peeling smears weights and reintroduces residual context. CLaP effectively captures a general rule in physical chemistry: addition of halogen groups increases the molecule's overall solubility in organic solvents and decreases the molecule's overall solubility in aqueous solvents. 

\subsection{FreeSolv Dataset}
We benchmark on the FreeSolv dataset, which contains \emph{642 small organic molecules} with experimentally measured \emph{hydration free energies in water} reported as

\[
\Delta G_{\mathrm{hyd}} \;[\mathrm{kcal/mol}].
\]

Thus, the sign directly encodes favorability: \textit{negative} values mean exergonic (favorable) hydration and greater water affinity, values near \(0\) indicate weak preference, and \textit{positive} values mean endergonic (unfavorable) hydration and greater hydrophobicity. \\

\begin{figure}[h]
  \centering
  \includegraphics[width=\linewidth]{\detokenize{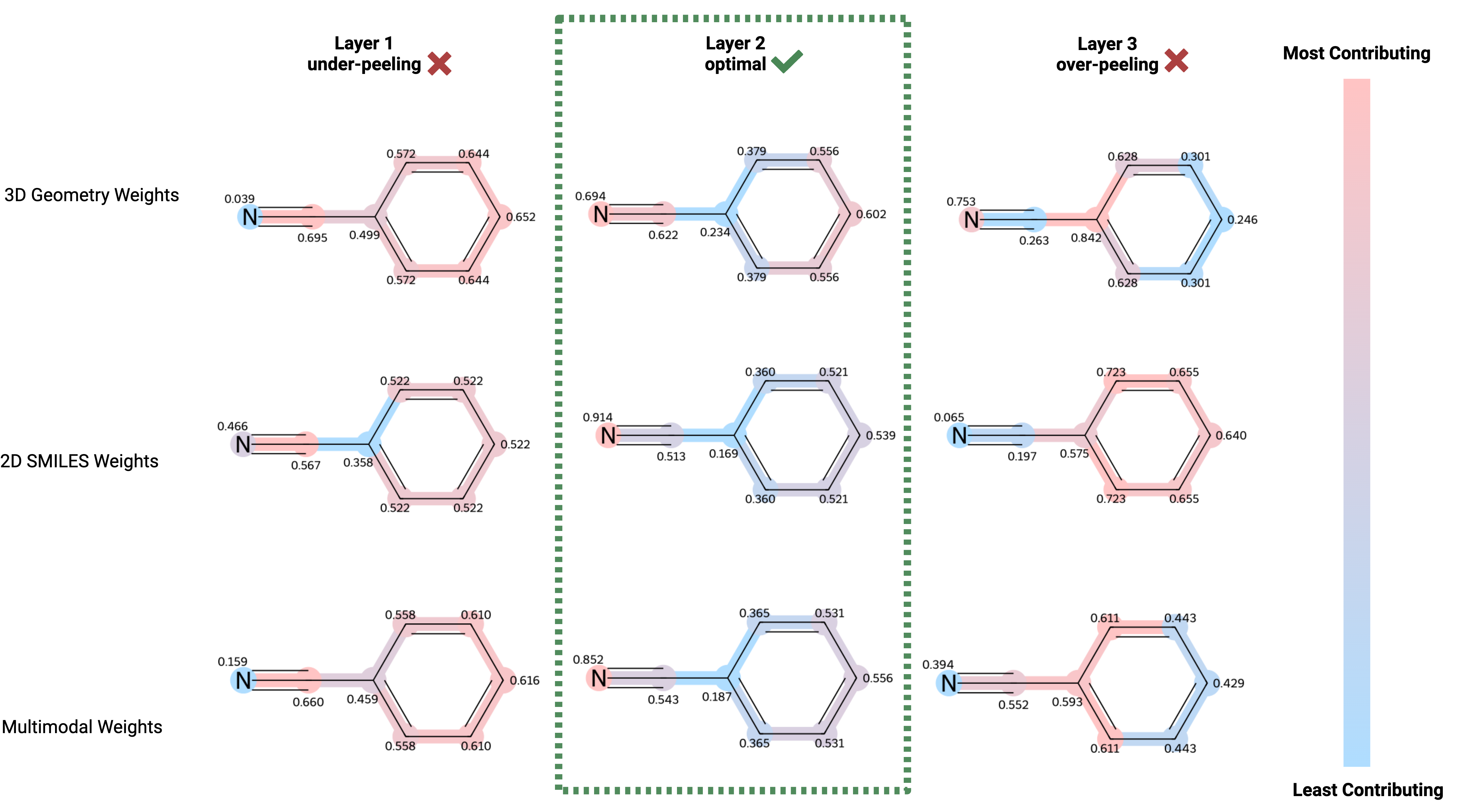}}
  \caption{Layer-wise attribution maps across input modalities. Red colored atoms are attributed with affecting the molecule's hydrophilicity the most.}
  \label{fig:freesolvap1}
\end{figure}

For Figure \ref{fig:freesolvap1}, at the optimal depth (L2), CLaP assigns the highest weights to the \emph{nitrile substituent} and the adjacent aromatic carbons; the 3D branch aggregates the linear C$\equiv$N dipole into a localized polar feature, while the 2D branch captures the nitrile group and ring context from connectivity. This depth-wise focus is exactly the intended ``peel'': batch-coupled context is removed, and label-relevant structure (a polar handle appended to an aryl
surface) is retained. The outcome matches physical chemistry: the nitrile increases polarity and provides a strong dipole acceptor, raising aqueous solubility relative to an unsubstituted aryl. Nitriles are widely used in medicinal chemistry contexts as metabolically stable polar groups that tune solubility and binding while maintaining a compact scaffold. \\

\begin{figure}[h]
  \centering
  \includegraphics[width=\linewidth]{\detokenize{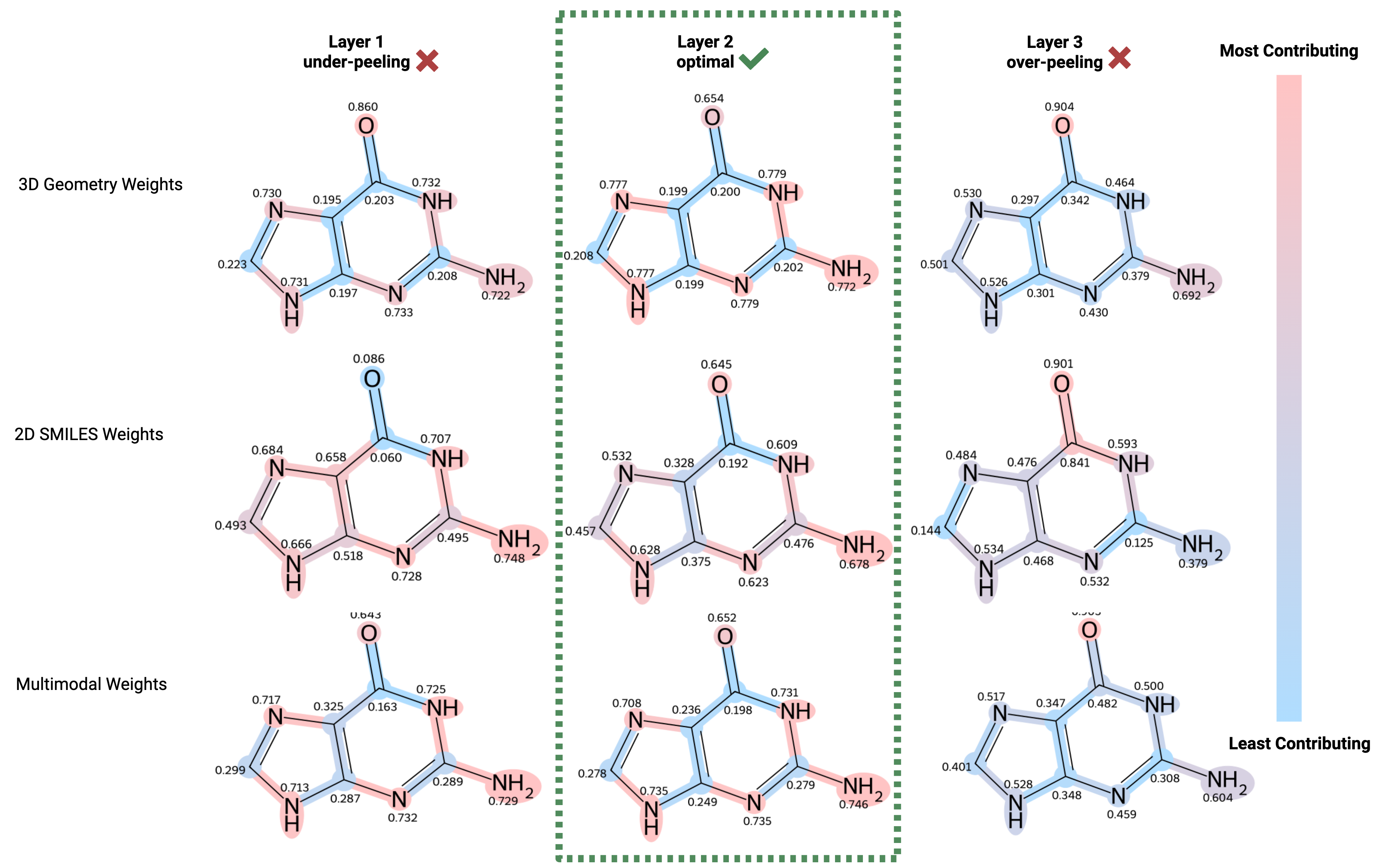}}
  \caption{Layer-wise attribution maps across input modalities. Red colored atoms are attributed with affecting the molecule's hydrophilicity the most.}
  \label{fig:freesolvap2}
\end{figure}

For Figure \ref{fig:freesolvap2}, at the optimal depth (L2), CLaP assigns the highest weights to the \emph{carbonyl oxygen} and the \emph{exocyclic amine group}, with additional contributions from the ring nitrogens. The 3D branch aggregates the localized dipoles into a polar spine, while the 2D branch captures the carbonyl and amino substituents from connectivity. This matches physical chemistry: the carbonyl oxygen and multiple NH$_2$/NH groups provide hydrogen-bond acceptors and donors, substantially raising aqueous solubility relative to a purely carbocyclic core. Under-peeling diffuses weights across the ring scaffold (shortcuts), while over-peeling smears attribution and reintroduces residual context. Heteroaryl amides of this type are commonly found in nucleobases and drug-like scaffolds, where their balance of hydrogen-bonding capacity and planarity helps tune solubility in chemical biology.

\subsection{Lipo Dataset}
We benchmark on the Lipophilicity (Lipo) dataset which contains
\emph{4,200 small molecules} curated experimentally. The target is the octanol/water distribution coefficient at physiological pH (\(\log D_{7.4}\)), defined as
\[
\log D_{7.4} \;\equiv\; \log_{10}\!\left(\frac{[C]_{\text{octanol}}}{[C]_{\text{water}}}\right)_{\mathrm{pH}=7.4},
\]
where concentrations include all ionization states at pH 7.4. Thus, \emph{higher (more positive)}
\(\log D_{7.4}\) 
denotes \emph{greater lipophilicity} (preference for octanol), whereas \emph{lower/negative} values indicate \emph{greater hydrophilicity}.

\begin{figure}[h]
  \centering
  \includegraphics[width=\linewidth]{\detokenize{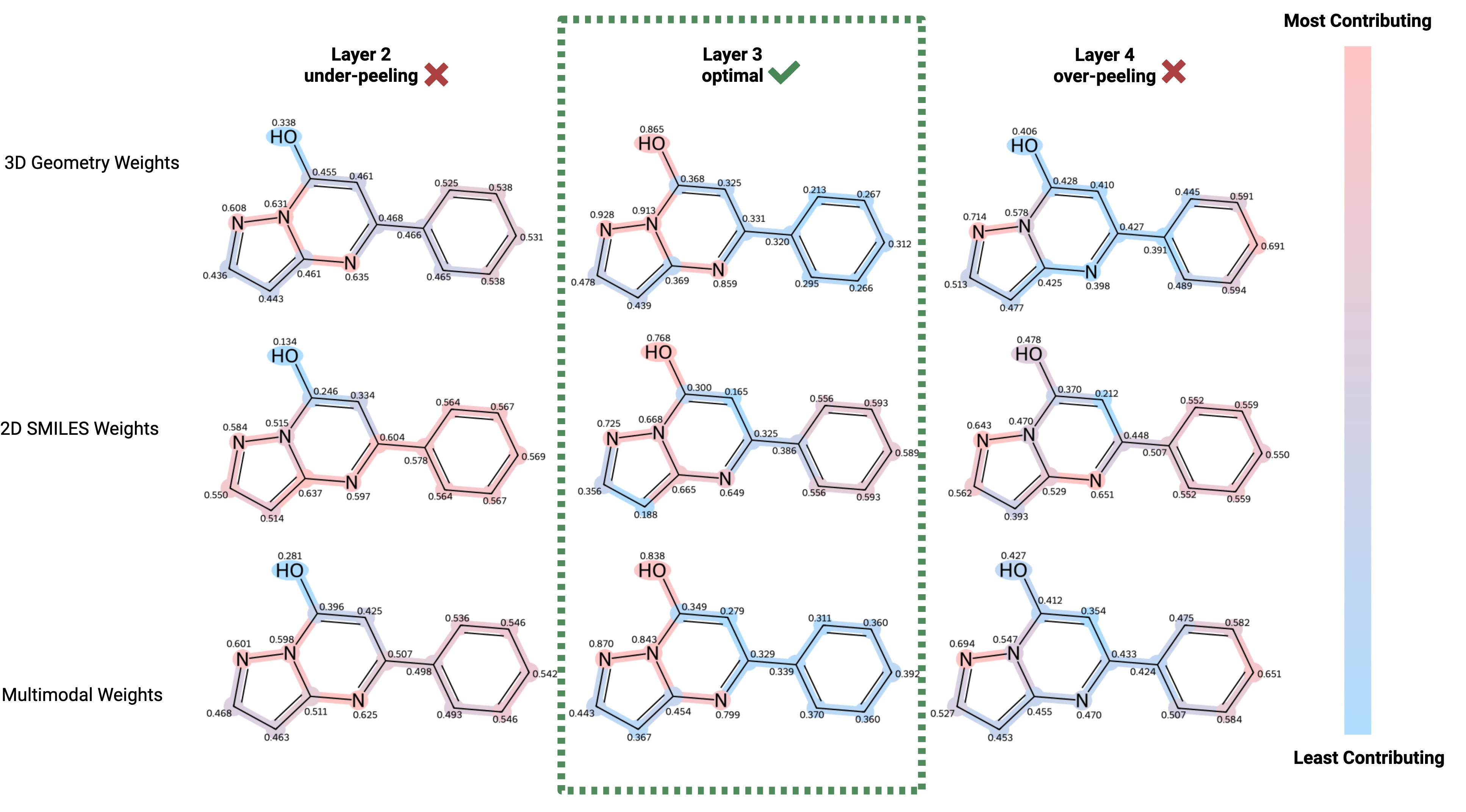}}
  \caption{Layer-wise attribution maps across input modalities. Red colored atoms are attributed with affecting the molecule's hydrophilicity the most.}
  \label{fig:lipoap1}
\end{figure}

For Figure \ref{fig:lipoap1}, at the optimal depth (L3), CLaP assigns the highest weights to the
\emph{hydroxyl substituent} and nearby heteroaryl nitrogens. The 3D branch aggregates the O–H dipole into a localized polar hotspot, while the 2D branch highlights the hydroxyl group and its attachment point to the heteroaromatic ring. This depth-wise focus is exactly the intended ``peel'': batch-coupled context is removed, and label-relevant structure (polar donors/acceptors distributed across a largely aromatic scaffold) is retained. The outcome matches physical chemistry: the hydroxyl provides a strong hydrogen-bond donor and acceptor site, partially counterbalancing the extended aromatic surface. As a result, the molecule is predicted to be more \emph{hydrophilic} than a fully unsubstituted aromatic analog. 

\begin{figure}[h]
  \centering
  \includegraphics[width=\linewidth]{\detokenize{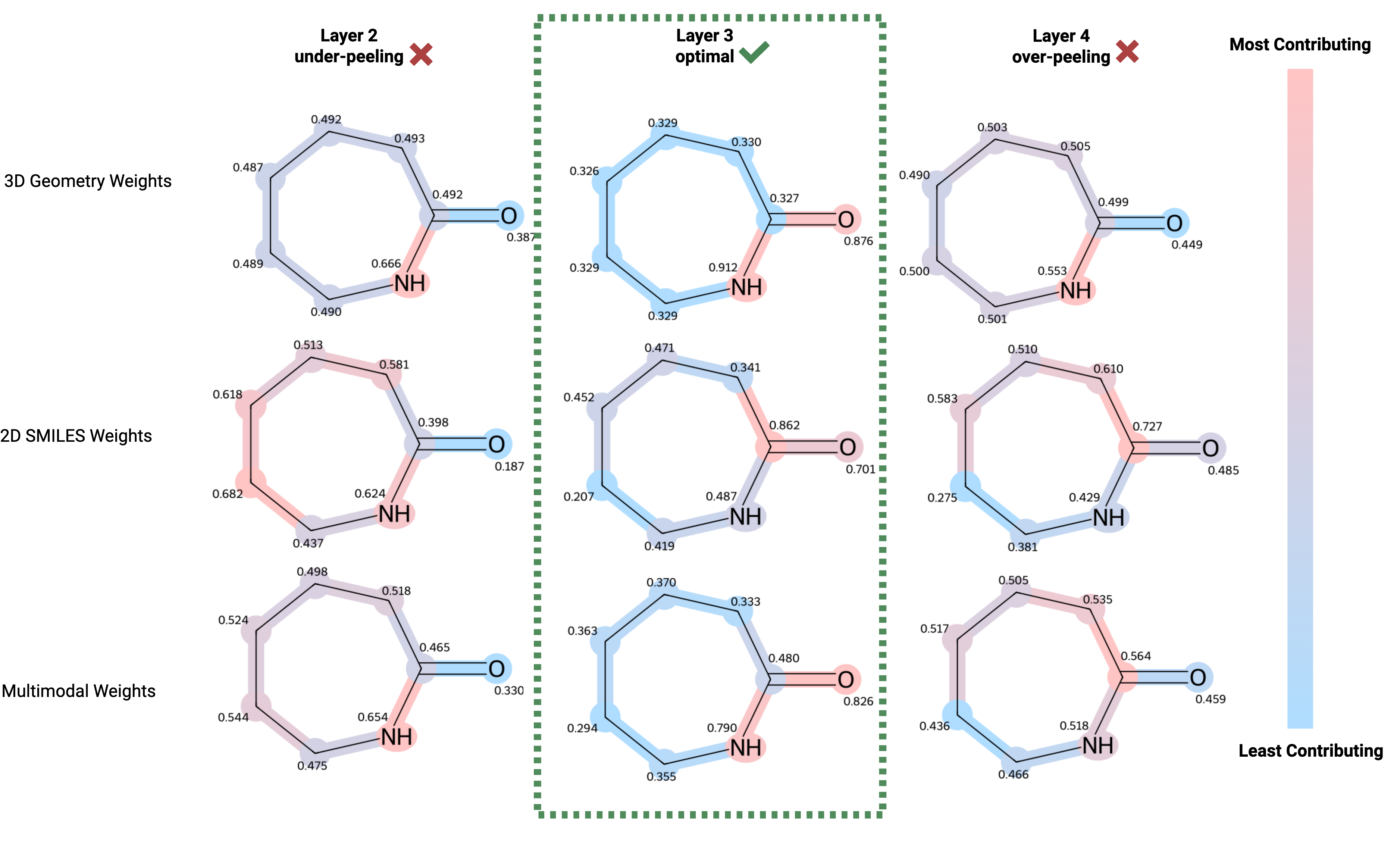}}
  \caption{Layer-wise attribution maps across input modalities. Red colored atoms are attributed with affecting the molecule's hydrophilicity the most.}
  \label{fig:lipapp2}
\end{figure}

For Figure \ref{fig:lipapp2}, at the optimal depth (L3), CLaP assigns the highest weight to the
\emph{amide carbonyl oxygen} and the nearby amide nitrogen. The 3D branch captures the strong local dipole between the C=O and NH, while the 2D branch highlights the amide linkage from connectivity. This depth-wise focus is exactly the intended ``peel'': batch-coupled context is removed, and label-relevant structure (a polar amide motif appended to a hydrophobic ring) is retained. CLaP captures a general modification in chemistry: the amide group generally introduces both a hydrogen-bond donor and acceptor, increasing polarity and reducing lipophilicity relative to an unsubstituted carbocyclic scaffold. As a result, the molecule is predicted to be more \emph{hydrophilic} (lower $\log D_{7.4}$).

\subsection{CycPeptMPDB Dataset}
We benchmark on the CycPeptMPDB dataset, which contains \emph{7,334 cyclic peptides} with experimentally measured \emph{passive membrane permeability} reported on a logarithmic scale as
\[
\log P \;\equiv\; \log_{10}\!\big(P\;[\mathrm{cm/s}]\big).
\]
Thus, higher (less negative) \(\log P\) denotes more permeable peptides, whereas more negative values indicate poor permeability.

\begin{figure}[h]
  \centering
  \includegraphics[width=\linewidth]{\detokenize{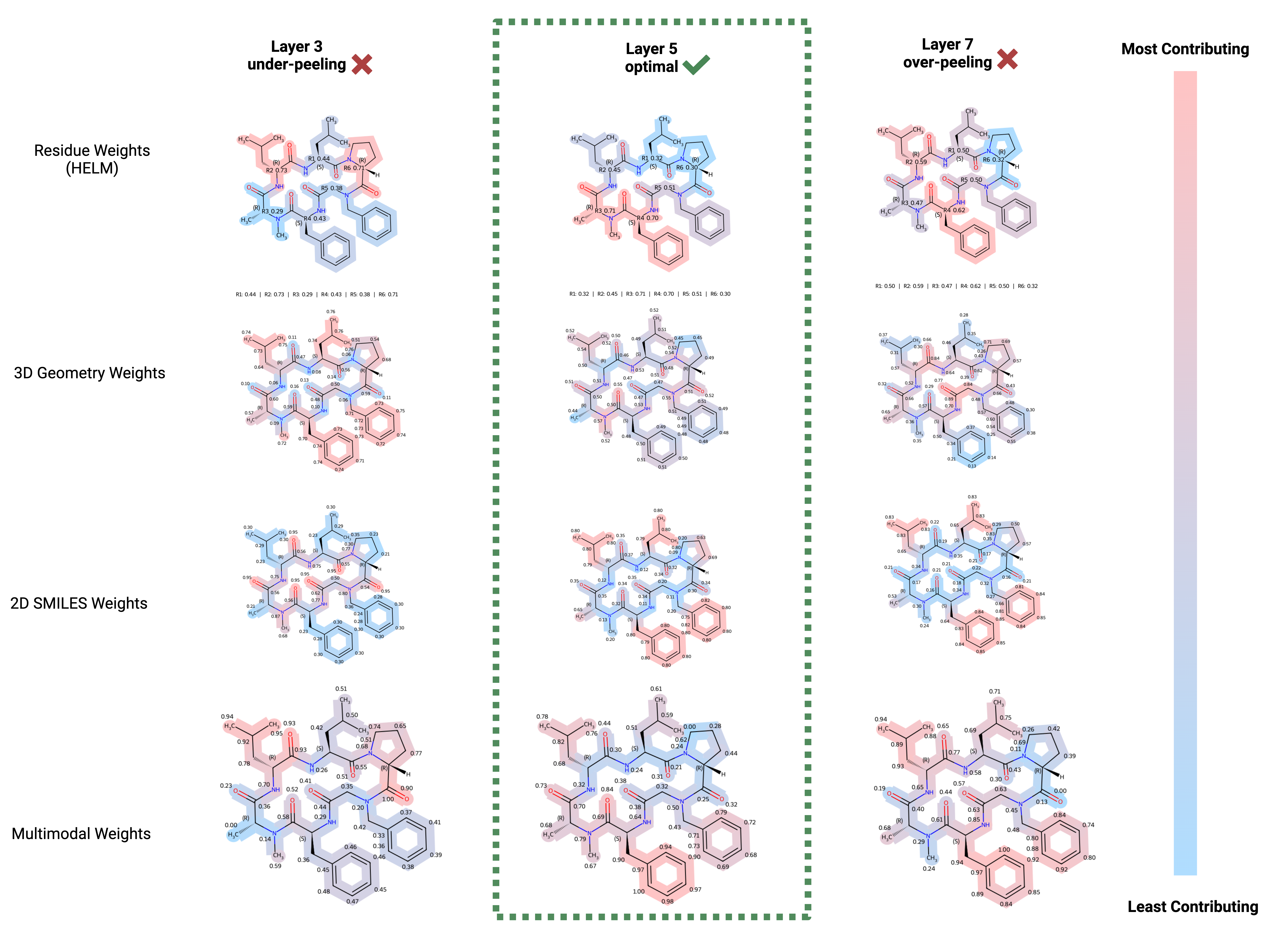}}
  \caption{Layer-wise attribution maps across input modalities. Red colored atoms are attributed with affecting the molecule's permeability the most.}
  \label{fig:pep1}
\end{figure}

For Figure \ref{fig:pep1}, at the optimal depth (L5), CLaP highlights \emph{hydrophobic residues} such as phenylalanine and aliphatic side chains as the dominant contributors. The 3D  branch aggregates bulky aromatic surfaces into hydrophobic patches, while the HELM representation captures residue-level context. This depth-wise focus is exactly the intended ``peel'': batch-coupled noise is removed, and label-relevant features (hydrophobic residues that enhance membrane passage) are retained. The outcome is consistent with physical chemistry and with what is known for peptide therapeutics: aromatic and aliphatic substitutions increase lipophilicity and improve passive \emph{cell permeability}. More broadly, such analyses underscore the drug-design logic of cyclic and linear peptide scaffolds: by balancing polarity (for solubility) with nonpolar residues (for permeability), peptide-based molecules can be optimized as potential drugs. Peptide drugs already play a major role in therapy,  ranging from hormones and enzyme inhibitors to next-generation modulators of protein–protein interactions.

\begin{figure}[h]
  \centering
  \includegraphics[width=\linewidth]{\detokenize{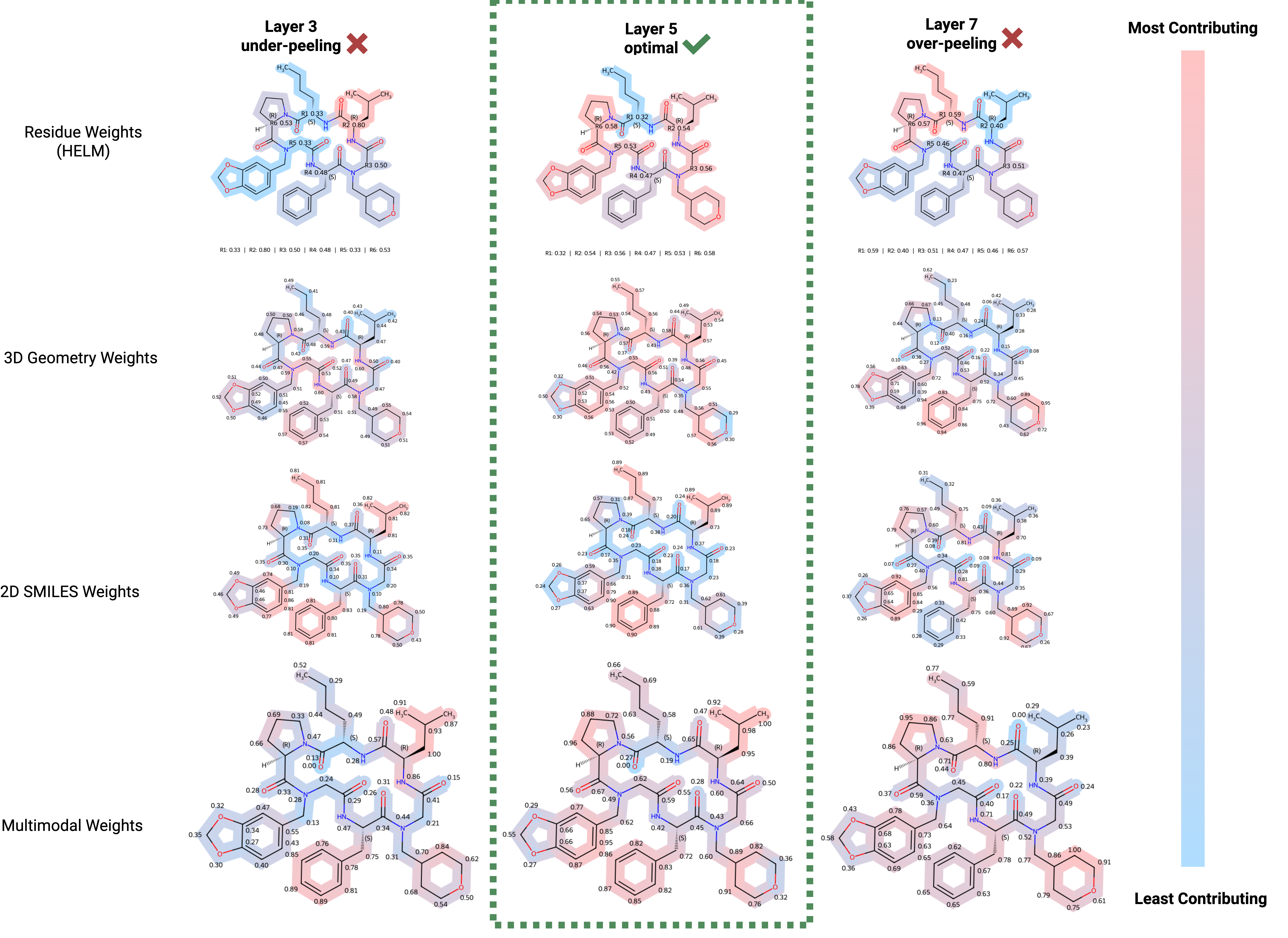}}
  \caption{Layer-wise attribution maps across input modalities. Red colored atoms are attributed with affecting the molecule's permeability the most.}
  \label{fig:pep2}
\end{figure}

For Figure ~\ref{fig:pep2}, at the optimal depth (L5), CLaP again highlights \emph{hydrophobic residues} such as aromatics and aliphatic side chains as the main drivers of permeability. These groups form contiguous lipophilic patches, favoring bilayer partitioning and 
consistent with design principles for permeable peptides. In contrast, \emph{heteroatom-rich sites} such as oxygen incorporated cycles contribute less, reflecting their polarity and the desolvation penalty that reduces passive diffusion. Thus, the model recovers a chemically intuitive balance: hydrophobic groups increase permeability, while oxygen and general incorporation of polar atoms in the side chains reduce permeability. CLaP is able to distinguish both drivers of permeability, as well as identify motifs that harm permeability, suggesting molecular edits that chemists can make to optimize the peptide. 

\section{Implementation Details}

\subsection{Proposed Framework Implementation Details}
\label{sec:implementation_details}

Our framework instantiates modality-specific encoders for 2D molecular graphs, sequence representations, and 3D geometries. Specifically, we encode 2D molecular graphs and HELM strings with GAT and 3D geometries with EGNN(3D structures apply ETKDG-up to 5 initial conformers, and energy-minimize the first embedded conformer using UFF); each backbone has 3 layers, producing 128-D embeddings that are linearly projected to a shared 256-D space. The target correlation schedule increases linearly from $0.5$ to $0.8$ across $L{=}5$ causal blocks. Training runs for 100 epochs with batch size 16 on $8\times$ RTX-6000 GPUs with early stopping. Loss weights are $\lambda_{\mathrm{caus}}{=}1.0$, $\lambda_{\mathrm{unif}}{=}1.0$, and $\lambda_{\mathrm{mono}}{=}0.5$; cross-modal consistency is disabled ($\lambda_{\mathrm{cons}}{=}0$), and the hinge margin is set to $0$. Our implementation is publicly available at an anonymous GitHub repository
(\textbf{\url{https://anonymous.4open.science/r/CLaP-FE64}}).

% For \emph{ESOL}, \emph{FreeSolv}, and \emph{Lipo} (non-peptide datasets), we use two modalities (2D graphs and 3D geometries); unless noted otherwise, embeddings are projected to 256-D and the correlation schedule is $0.4{\rightarrow}0.7$. ESOL uses width 128 with 2 encoder layers and $L{=}2$ causal blocks; FreeSolv uses width 64 with 2 layers and $L{=}2$; Lipo uses width 64 with 3 layers and $L{=}3$. The hinge margin is $0$ for all settings.

\subsection{Baselines Implementation Details}
\label{sec:baseline_implementation_details}

For causality-oriented baselines, we modified the original classification-based causal framework into regression models for molecular property prediction. Specifically, the classification head was replaced with a regression head by fixing the output layer dimension to one. At the data level, we replaced the original discrete datasets with SMILES-derived PyG graph structures labeled with continuous molecular properties (e.g., permeability, logD74). For evaluation, metrics were changed from accuracy/AUC to MSE, RMSE, MAE, and R². These modifications preserve the structural features of the original architecture while adapting it to our new regression tasks.
% \subsection{The Use of Large Language Models (LLMs)}
% \label{sec:useofLLMs}
% We used an LLM solely for light editorial assistance (grammar, wording, and minor style). The LLM did not contribute ideas, methods, models, experiments, analyses, or decisions, and was not used to generate or label data or code. All scientific content is by the authors.

\end{document}